\documentclass[10pt,journal,compsoc]{IEEEtran}
\ifCLASSOPTIONcompsoc
\usepackage{color}
\usepackage[ruled,vlined]{algorithm2e}
\usepackage{graphicx}
\usepackage{subfigure}
\usepackage{enumitem}
\usepackage{multirow}
\usepackage{url}
\usepackage{amsthm,amsmath,amssymb}
\newcommand{\sen}[1]{{\color{black}{#1}}}

\newcommand{\revise}[1]{{\color{black}{#1}}}
\usepackage{pdfpages}
\newtheorem{definition}{Definition}
\newtheorem{theorem}{Theorem}
\newtheorem{proposition}{Proposition}

\usepackage[nocompress]{cite}
\else
  \usepackage{cite}
\fi

\ifCLASSINFOpdf
\else
\fi
\usepackage{stfloats}

\begin{document}

\title{Bipartite Ranking Fairness through a Model Agnostic Ordering Adjustment}
\author{Sen~Cui,
        Weishen~Pan,
        Changshui~Zhang,~\IEEEmembership{Fellow,~IEEE},
        Fei~Wang
\IEEEcompsocitemizethanks{
  \IEEEcompsocthanksitem S. Cui, W. Pan and C. Zhang are with the Department of Automation, BNRist, Tsinghua University, Beijing, China. F. Wang is with the Department of Population Health Sciences, Weill Cornell Medicine, Cornell University, New York, USA. Email: \{cuis19, pws15\}@mails.tsinghua.edu.cn, zcs@tsinghua.edu.cn, few2001@med.cornell.edu. \\
  Corresponding authors: Changshui Zhang and Fei Wang.}}

\markboth{Submitted to IEEE Transactions on Pattern Analysis and Machine Intelligence}%
{Shell \MakeLowercase{\textit{et al.}}: Bare Advanced Demo of IEEEtran.cls for IEEE Computer Society Journals}

\IEEEtitleabstractindextext{%
\begin{abstract}
Recently, with the applications of algorithms in various risky scenarios, algorithmic fairness has been a serious concern and received lots of interest in machine learning community. In this paper, we focus on the bipartite ranking scenario, where the instances come from either the positive or negative class and the goal is to learn a ranking function that ranks positive instances higher than negative ones. We are interested in whether the learned ranking function can cause systematic disparity across different protected groups defined by sensitive attributes. While there could be a trade-off between fairness and performance, we propose a model agnostic post-processing framework \texttt{xOrder} for achieving fairness in bipartite ranking and maintaining the algorithm classification performance. In particular, we optimize a weighted sum of the utility as identifying an optimal warping path across different protected groups and solve it through a dynamic programming process. \texttt{xOrder} is compatible with various classification models and ranking fairness metrics, including supervised and unsupervised fairness metrics. In addition to binary groups, \texttt{xOrder} can be applied to multiple protected groups. We evaluate our proposed algorithm on four benchmark data sets and two real-world patient electronic health record repositories. \texttt{xOrder} consistently achieves a better balance between the algorithm utility and ranking fairness on a variety of datasets with different metrics. From the visualization of the calibrated ranking scores, \texttt{xOrder} mitigates the score distribution shifts of different groups compared with baselines. Moreover, additional analytical results verify that \texttt{xOrder} achieves a robust performance when faced with fewer samples and a bigger difference between training and testing ranking score distributions. For reproducing our results, we also made our source codes available at \url{https://github.com/cuis15/xorder}.
\end{abstract}

% Note that keywords are not normally used for peer review papers.
}
\maketitle
\IEEEdisplaynontitleabstractindextext
\IEEEpeerreviewmaketitle
\section{Introduction}
These days, machine learning algorithms have been widely applied in a variety of real-world applications~\cite{he2016deep,vaswani2017attention} including the high-stakes scenarios~\cite{park2019tackling} such as loan approvals, criminal justice, healthcare~\cite{nie2014bridging}, etc. An increasing concern is whether these algorithms make fair decisions in these cases. For example, ProPublica reported that an algorithm used across the US for predicting a defendant’s risk of future crime produced higher scores to African-Americans than Caucasians on average~\cite{angwin2016machine}. This stimulates lots of research on improving the fairness of the decisions made by machine learning algorithms.

When mitigating the disparity informed by algorithms, existing works have mostly focused on the disparate impacts of binary decisions~\cite{dwork2012fairness,hardt2016equality,zehlike2017fa,zhang2020fairness} with respect to different groups formed from the protected variables (e.g., gender or race). Various fairness metrics have been proposed to quantify the algorithmic disparity in diverse situations, including unsupervised and supervised criteria. \revise{For example, demographic parity~\cite{kamiran2012data,dwork2012fairness} is an unsupervised fairness metric that measures whether the classification results are independent of the group membership.} Equalized odds~\cite{hardt2016equality} seeks for equal false positive and negative rates across different groups. Accuracy parity~\cite{zafar2017fairness} needs equalized error rates across different groups.

In addition to the binary classification, another scenario that frequently involves computational algorithms is ranking~\cite{liu2009learning,adomavicius2011improving,haveliwala2003topic,long2014active}. For example, Model for End-stage Liver Disease (MELD) score, which is derived from a simple linear model from several features, has been used for prioritizing candidates who need liver transplantation~\cite{wiesner2003model}. \revise{Studies have shown that the use of MELD score as a ranking system can lead to gender bias, as women are less likely to receive a liver transplant within three years compared to men~\cite{moylan2008disparities}.} Extensive research has been studied to quantify the ranking fairness~\cite{kallus2019fairness,beutel2019fairness,narasimhan2020pairwise,kuhlman2019fare,vogel2020learning}. For example, Kallus {\em et al.}~\cite{kallus2019fairness} proposed xAUC, which measures the probability of positive examples of one group being ranked above negative examples of another group. Beutel {\em et al.}~\cite{beutel2019fairness} proposed a similar definition \emph{pairwise ranking fairness} (PRF), which requires equal probabilities for positive instances from each group ranked above all negative instances. Vogel \emph{et al.}~\cite{vogel2020learning} give a stricter criterion that requires the equality between distributions from different groups.
\begin{figure}[h]
    \centering
    \includegraphics[width=0.95\columnwidth]{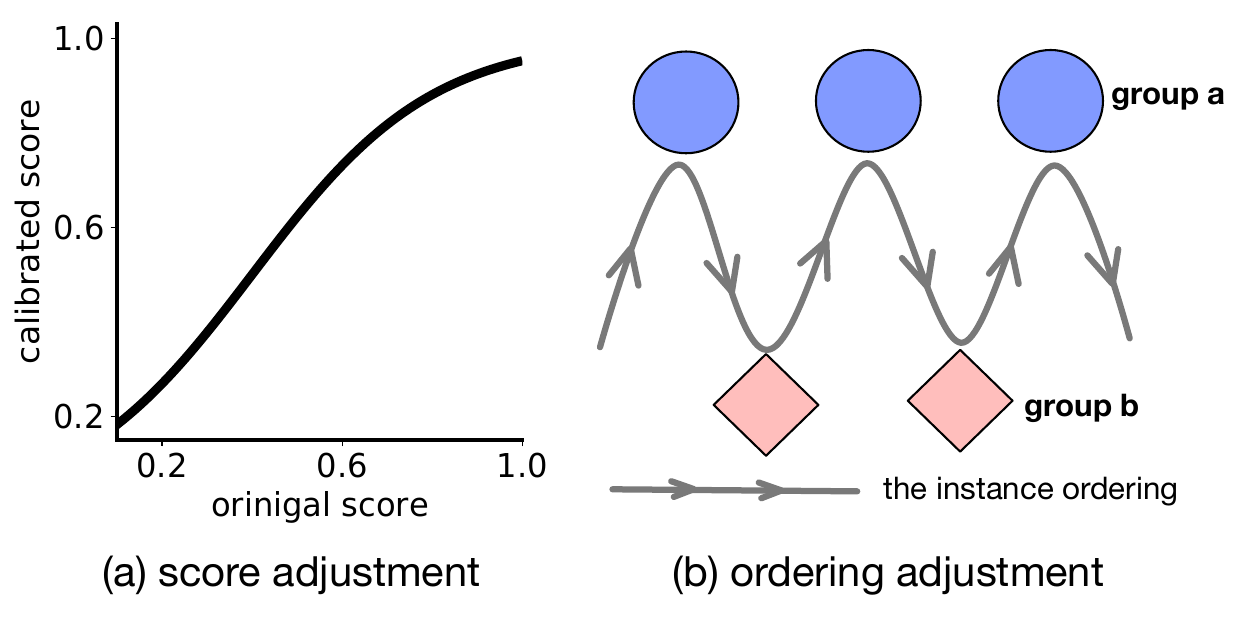}
    \caption{Illustrations of score adjustment and ordering adjustment. (a). the transformation function for adjusting ranking scores to optimize the ordering; (b) \texttt{xOrder} adjusts the instance ordering directly.}
    \label{fig:intro}
\end{figure}

A variety of efforts have been made to address the potential disparity induced by risk scores. A recent line of research proposes to optimize an objective with a fairness regularization to calibrate the score distribution. Beutel {\em et al.}~\cite{beutel2019fairness} studied the balance between algorithm utility and ranking fairness, and proposed an optimization framework by minimizing an objective including the classification loss and a regularization term evaluating the absolute correlation between the group membership and pairwise residual predictions. Though this method considers both utility and fairness, is model-dependent and does not directly optimize the disparity but an approximated proxy. Other research proposes to learn a post-processing function to calibrate the score distribution. For example, Kallus {\em et al.}~\cite{kallus2019fairness} propose to adjust the risk scores with a parameterized monotonically increasing function as shown in Figure~\ref{fig:intro}. This method is model agnostic and aims to achieve equal xAUC, but it does not consider algorithm utility (i.e., classification performance) explicitly.

Inspired by the generality of the post-processing method, in this paper, we develop \texttt{xOrder} to achieve ranking fairness and maintain the algorithm utility simultaneously, in a more flexible way. Specifically, we show that both algorithm utility and ranking fairness are essentially determined by the ordering of the instances involved. \texttt{xOrder} makes direct adjustments of the cross-group instance ordering (while existing post-processing algorithms mostly aimed at adjusting the ranking scores to optimize the ordering) as shown in Figure~\ref{fig:intro}. The optimal adjustments can be obtained through a dynamic programming procedure of minimizing an objective comprising a weighted sum of algorithm utility loss and ranking disparity. We theoretically analyze our method in two cases. If we focus on maximizing the utility, \texttt{xOrder} achieves a globally optimal solution. While we care only about minimizing the disparity, it can have a relatively low bound of ranking disparity. The learned ordering adjustment can be easily transferred to the test data through interpolation methods.

To verify the effectiveness of our proposed method, we evaluate \texttt{xOrder} on four popular benchmark data sets for studying algorithm fairness and two real-world electronic health record data repositories. The results show \texttt{xOrder} achieves low ranking disparities on all data sets while at the same time maintaining good algorithm utilities. In addition, we compare the performance of \texttt{xOrder} with another post-processing algorithm when faced with the difference between training and test distributions. From the results, our algorithm can achieve robust performance when training and test ranking score distributions are significantly different. The source codes of \texttt{xOrder} are made publicly available at \url{https://github.com/cuis15/xorder}.

Part of the results is published originally in~\cite{cui2021towards} at SIGKDD 2021. It is worthwhile to highlight our substantial extensions in several aspects:

\begin{itemize}
  \item we consider a more general ranking fairness problem where the disparity could be quantified by various unsupervised or supervised metrics in a unified formulation;

  \item besides binary groups, we provide a general disparity defined across multiple groups and advance our method to a multi-dimensional \texttt{xOrder} to mitigate this disparity; we also propose a more efficient variant of \texttt{xOrder} for multi-group ranking fairness problems;

  \item we conduct comprehensive experiments on a series of settings to verify the validity of our method, including the scenarios with more data samples and protected groups. The experiments show that our method could achieve a better trade-off between fairness and utility compared with baselines on diverse scenarios;

  \item to intuitively illustrate the effect of our method, we show the visualization of the learned scores before and after the adjustments; in addition to minimizing the gap of AUC among groups, our method encourages the equality of the score distributions;

  \item more theoretical and analytical results are presented, including the upper bound of the disparity, analysis of the unfairness when there are multiple groups, etc.

\end{itemize}

\revise{The subsequent sections of the paper are organized as follows: Section 2 provides a brief overview of the relevant literature in the field. Section 3 establishes the necessary notation, definitions, and problem formulation for achieving fairness in bipartite ranking. In Section 4, we present the details of our proposed framework, \texttt{xOrder}, and demonstrate its ability to balance fairness and utility. The effectiveness of our method is further validated through experimental results in Section 5, where we compare it against baseline methods. Section 6 examines the robustness of our approach in scenarios involving limited training data and significant differences between training and test distributions. Finally, we conclude the paper in Section 7.}

% The rest of this paper is organized as follows. In Section~\ref{sec:2}, we look back on previous work in algorithmic fairness, especially the research about bipartite ranking fairness. In Section~\ref{sec:3}, we introduce the notations and formulate the problem for balancing ranking fairness and performance. In Section~\ref{sec:4}, we propose our framework \texttt{xOrder} in detail. In section~\ref{sec:5} and~\ref{sec:6}, we experimentally validate the effectiveness and reliability of our proposed method. Finally, Section~\ref{sec:6} draws our conclusions.

% \sen{In addition, we assess the robustness of our algorithm in two aspects. We conduct a series of experiments given fewer training samples and \texttt{xOrder} maintains its advantages. We also compare the performance with another post-processing algorithm when faced with the difference between training and test distributions. From the results, we find it can achieve robust performance when training and test ranking score distributions are significantly different.}

\section{Related Works}
\label{sec:2}
\textbf{Algorithmic Fairness} Algorithm fairness is defined as the disparities in the decisions made across groups formed by protected variables, such as gender and race. Many previous works on this topic focused on binary decision settings. Researchers have used different proxies as fairness measures which are required to be the same across different groups for achieving fairness. Examples of such proxies include the proportion of examples classified as positive~\cite{calders2009building,calders2010three}, as well as the prediction performance metrics such as true/false positive rates and error rates~\cite{dixon2018measuring,feldman2015certifying,hardt2016equality,zafar2017fairness,kallus2018residual}. A related concept that is worthy of mentioning here is calibration~\cite{lichtenstein1981calibration}. A model with risk score $\operatorname{S}$ on input $\operatorname{X}$ to generate output $\operatorname{Y}$ is considered calibrated by group if for $\forall s\in\left\{0,1\right\}$, we have $\operatorname{Pr}(\operatorname{Y} = 1| \operatorname{S} = s, \operatorname{A} = a) = \operatorname{Pr}(\operatorname{Y} = 1| \operatorname{S} = s, \operatorname{A} = b)$ where $\mathrm{A}$ is the group variable~\cite{chouldechova2017fair}. Recent studies have shown that it is impossible to satisfy both error rate fairness and calibration simultaneously when the prevalence of positive instances are different across groups~\cite{kleinberg2016inherent,chouldechova2017fair}. Plenty of approaches have been proposed to achieve fairness in binary classification settings. One type of method is to train a classifier without any adjustments and then post-process the prediction scores by setting different thresholds for different groups~\cite{hardt2016equality}. Other methods have been developed for the optimization of fairness metrics during the model training process through adversarial learning~\cite{zemel2013learning,louizos2015variational,beutel2017data,madras2018learning,zhang2018mitigating} or regularization~\cite{kamishima2011fairness,zafar2015fairness,beutel2019putting}.

\noindent \textbf{Ranking fairness} Ranking fairness is an important issue in applications where the decisions are made by algorithm-produced ranking scores, such as the example of liver transplantation candidate prioritization with MELD score~\cite{wiesner2003model}. This problem is related to but different from binary decision making~\cite{narasimhan2013relationship,menon2016bipartite}. There are prior works formulating this problem in the setting of selecting the top-k items ranked based on the ranking scores for any k~\cite{celis2017ranking,yang2017measuring,zehlike2017fa,geyik2019fairness}. For each sub-problem with a specific k, the top-k ranked examples can be treated as positive while the remaining examples can be treated as negative, so that these sub-problems can be viewed as binary classification problems. There are works trying to assign a weight to each instance according to the orders and study the difference of such weights across different groups~\cite{singh2018fairness,singh2019policy}. Our focus is the fairness on bipartite ranking, which seeks a good ranking function that ranks positive instances above negative ones~\cite{menon2016bipartite}. Kallus {\em et al}.~\cite{kallus2019fairness} defined xAUC (Area under Cross-Receiver Operating Characteristic curve) as the probability of positive examples of one group being ranked above negative examples of another group. They require equal xAUC to achieve ranking fairness. Beutel {\em et al.}~\cite{beutel2019fairness} proposed a similar definition of pairwise ranking fairness(PRF) as the probability that positive examples from one group are ranked above all negative examples and use the difference of PRF across groups as a ranking fairness metric. They further proved that some traditional fairness metrics (such as calibration and MSE) are insufficient for guaranteeing ranking fairness under PRF metric. Vogel \emph{et al.}~\cite{vogel2020learning} argue that the AUC-constrained criteria cannot guarantee the equality of the score distributions and propose a stricter criterion that encourages similar distributions from different groups.

Previous methods mainly focus on calibrating the risk scores instead of adjusting the ordering directly. For example, Kallus {\em et al}.~\cite{kallus2019fairness} proposed a post-processing technique. They transformed the prediction scores in the disadvantaged group with a logistic function and optimized the empirical xAUC disparity by exhaustive searching on the space of parameters without considering the trade-off between algorithm utility and fairness. As the objective of the ranking problem is non-differentiable, there are theoretical and empirical works that propose to apply a differentiable objective to approximate the original non-differentiable objective~\cite{vogel2020learning,narasimhan2020pairwise,beutel2019putting}. Vogel {\em et al}. propose to use a logistic function as smooth surrogate relaxations and provide upper bounds of the difference between the global optima of the original and the relaxed objectives. Narasimhan {\em et al.}~\cite{narasimhan2020pairwise} reduced ranking problems to constrained optimization problems and proposed to solve the problems by applying an existing optimization framework proposed in ~\cite{cotter2019two}. Beutel {\em et al.} proposed a pairwise regularization for the objective function~\cite{beutel2019fairness}. However, it is difficult to apply this regularization to some learning methods such as boosting model~\cite{freund2003efficient}, since it is challenging to reweight the samples during the boosting iterations, because the impact of increasing/decreasing the weight of the samples on fairness is difficult to control. Moreover, the method in~\cite{beutel2019fairness} considers only binary protected groups, which may limit its applications when there are multiple groups.
\begin{figure*}[h]
    \centering
    \includegraphics[width=1.95\columnwidth]{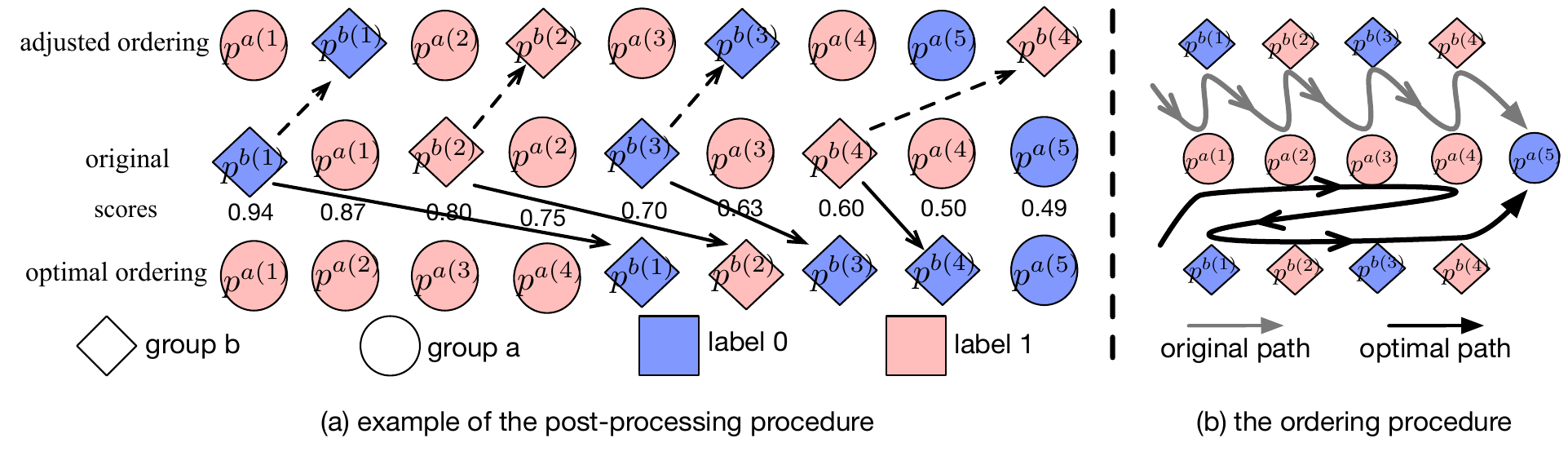}
    \caption{(a).An example to illustrate the post-processing. Original scores are in the middle. The first row is the ordering after post-processing, while the optimal ordering is on the bottom; (b).Illustrations of the ordering procedure as path finding for the example in (a).}
    \label{fig:samples}
\end{figure*}
\section{Bipartite Ranking Fairness Problem}
\label{sec:3}
\subsection{Notations}
Suppose we have data ($\operatorname{X}$, $\operatorname{A}$, $\operatorname{Y}$) on features $\operatorname{X} \in \mathcal{X}$, sensitive attribute $\operatorname{A} \in \mathcal{A}$ and binary label $\operatorname{Y} \in\{0,1\}$. We are interested in the performance and the fairness issue of a predictive ranking score function $R$: $\mathcal{X} \times \mathcal{A} \rightarrow \mathbb{R}$. Here we focus on the case where $R$ returns an estimated conditional probability positive label corresponding to a given individual's ranking score. We use $ \mathrm{S}=R(\operatorname{X}, \operatorname{A})\in[0,1]$ to denote the individual ranking score variable and the score $\mathrm{S}$ means the probability that the individual belongs to the positive group.

Given $\operatorname{S}$, we can derive a binary classifier with a given threshold $\theta$, such that $\hat{\operatorname{Y}}_{\theta}=\mathbb{I}[ \operatorname{S} \geq \theta]$ and $\mathbb{I}$ is the indicator function.

\noindent \textbf{Utility} To evaluate the performance of $R$, the receiver operator characteristic (ROC) curve is widely adopted with the false positive rate (FPR) on the x-axis and the true positive rate (TPR) on the y-axis as the threshold $\theta$ varies. The area under the ROC curve (AUC) quantitatively measures the quality of the learned scoring function $R$.

AUC can also be understood as the probability that a randomly drawn ranking score from the positive class is ranked above a randomly drawn score from the negative class~\cite{hanley1982meaning}

\begin{equation}
\label{eq:AUC}
\begin{aligned}
\mathrm{AUC} &= \operatorname{Pr}[\operatorname{S}_{1}>\operatorname{S}_{0}]\\
&= \frac{1}{n_{1} n_{0}} \cdot \sum\nolimits_{i:\mathrm{Y}_i=1}\sum\nolimits_{j:\mathrm{Y}_j=0}\mathbb{I}\left[R(\operatorname{X}_{i})>R(\operatorname{X}_{j})\right],
\end{aligned}
\end{equation}

where $\operatorname{S}_1$ and $\operatorname{S}_0$ represent a ranking score of a random positive and negative sample. $n_1$ and $n_0$ correspond to the number of positives and negatives, respectively. Note that we dropped the group variable in the $R$ function because it is irrelevant to the measure of AUC (i.e., $\mathrm{X}_i$, $\mathrm{X}_i$ can be from any group).

\subsection{Disparity Metrics with Two Protected Groups}
\label{sec:3-2}
Suppose there are two groups denoted as $a$ and $b$ formed by the sensitive variable $\mathrm{A}$. We construct an unsupervised fairness metric analog of the demographic parity~\cite{dwork2012fairness} metric.
\begin{definition}[Unsupervised Ranking Fairness (URF)] The URF for two protected the group a and the group b is defined as
\begin{equation}
\mathrm{URF}(a,b) = \left|\mathrm{Pr}[\mathrm{S}^{a} > \mathrm{S}^{b}] - \mathrm{Pr}[\mathrm{S}^{b} > \mathrm{S}^{a}] \right|,
\label{eq:urf}
\end{equation}
where $\operatorname{S}^{a}$ is the ranking score of a random sample in $a$. $\operatorname{S}^{b}$ is the ranking score of a random sample in $b$.
\label{def:urf}
\end{definition}
From Definition~\ref{def:urf}, $\mathrm{URF}$ requires that the two groups should have the same probability to rank above.

While $\mathrm{URF}$ ignores the possible difference of the distributions of the two groups, Kallus \emph{et al.}~\cite{kallus2019fairness} argue that the two group-level ranking fairness metrics can be measured by the Cross-Area Under the Curve (xAUC) metric, which denotes the cross-area under the cross-ROC (xROC) curve defined as follows.

\begin{definition}[Cross-Receiver Operating Characteristic curve (xROC)~\cite{kallus2019fairness}]
\begin{equation}
\mathrm{xROC}(\theta; \mathrm{S}, a, b)=(\operatorname{Pr}[\mathrm{S}^{b}_{0} >\theta], \ \operatorname{Pr}[\mathrm{S}^{a}_{1}>\theta]),
\label{eq:xroc}
\end{equation}
where $\mathrm{S}^{b}_{0}$ is the ranking score of a negative random sample in $b$. $\operatorname{S}^{a}_{1}$ is the ranking score of a random positive sample in $a$.
\end{definition}
The xROC curve presents $\mathrm{xROC}(\theta; \mathrm{S}, a, b)$ over the space of thresholds $\theta \in \mathbb{R}$, in which the curve of TPR of the group a is on the y-axis and the FPR of group b is on the x-axis.
\begin{definition}[xAUC~\cite{kallus2019fairness}] The xAUC of group $a$ over $b$ is defined as
\begin{equation}
\begin{aligned}
\label{eq:xAUC}
&\mathrm{xAUC}(a, b)=\operatorname{Pr}\left[\operatorname{S}_{1}^{a}>\operatorname{S}_{0}^{b}\right]\\
&= \frac{1}{n^{a}_{1} n^{b}_{0}} \sum_{i:i\in a,\mathrm{Y}_{i}=1}\sum_{j: j\in b, \mathrm{Y}_{j}=0}\mathbb{I} \left[R(\operatorname{X}_{i},a) > R(\operatorname{X}_{j},b)\right],
\end{aligned}
\end{equation}
$n^a_1$ and $n^b_0$ correspond to the number of positives in a and negatives in b, respectively. $i$ is the index of a particular positive sample from group $a$, whose corresponding ranking score is $R(\operatorname{X}_{i},a)$. $j$ is the index of a particular negative sample from group $b$, whose corresponding ranking score is $R(\operatorname{X}_{j},b)$.
\end{definition}
From Eq.(\ref{eq:xAUC}) we can see that xAUC measures the probability of a random positive sample in $a$ ranked higher than a random negative sample in $b$. Correspondingly, xAUC($b$,$a$) means $\operatorname{Pr}(\operatorname{S}_{1}^{b}>\operatorname{S}_{0}^{a})$, and the ranking disparity can be measured by
\begin{equation}
\label{eq:delta_xAUC}
\begin{aligned}
\Delta \mathrm{xAUC}(a,b) &=\left|\operatorname{xAUC}(a,b)-\operatorname{xAUC}(b,a)\right|\\
&=\left|\operatorname{Pr}[\operatorname{S}_{1}^{a}>\operatorname{S}_{0}^{b}]-\operatorname{Pr}[\operatorname{S}_{1}^{b}>\operatorname{S}_{0}^{a}]\right|.
\end{aligned}
\end{equation}
\begin{definition}[Pairwise Ranking Fairness (PRF)~\cite{beutel2019fairness}] The PRF for group $a$ is defined as
\begin{equation}
\label{eq:pairwise_ranking}
\begin{aligned}
\mathrm{PRF}(a) &=\operatorname{Pr}[ \operatorname{S}_{1}^{a}>\operatorname{S}_{0}]\\
&=\frac{1}{n^{a}_{1} \cdot n_{0}} \sum_{i:i\in a,\mathrm{Y}_{i}=1 }\sum_{j:\mathrm{Y}_{j}=0}\mathbb{I} \left[R(\operatorname{X}_{i},a) > R(\operatorname{X}_{j})\right],
\end{aligned}
\end{equation}
where sample $j$ can belong to either group $a$ or group $b$.
\end{definition}
From Eq.(\ref{eq:pairwise_ranking}) we can see that the PRF for group $a$ measures the probability of a random positive sample in $a$ ranked higher than a random negative sample in either $a$ or $b$. Then we can also define the following $\Delta$PRF metric to measure the ranking disparity
\begin{equation}
\label{eq:deltaPRF}
\Delta\mathrm{PRF}(a,b)=\left|\operatorname{Pr}[ \operatorname{S}_{1}^{a}>\operatorname{S}_{0}] - \operatorname{Pr}[\operatorname{S}_{1}^{b}>\operatorname{S}_{0}]\right|.
\end{equation}
\subsection{Disparity Metrics with Multiple Protected Groups}
In Section~\ref{sec:3-2}, we introduce an unsupervised fairness metric ($\Delta$URF) and two classic supervised metrics ($\Delta$xAUC and $\Delta$PRF) when there are two protected groups. When there are multiple protected groups, the disparity can be defined in a pairwise way. For example,
\begin{equation}
\Delta \mathrm{URF}(a_{1},...,a_{N}) = \max(\Delta\mathrm{URF}(a_{i},a_{j})), \ i, j \in [N], i\not=j.
\label{eq:mrf}
\end{equation}

\noindent \textbf{No dominated group.} Suppose we define the group $a_{i}$ is the advantaged group if $\mathrm{Pr}[\mathrm{S}^{a_{i}} > \mathrm{S}^{a_{j}}] > \mathrm{Pr}[\mathrm{S}^{a_{j}} > \mathrm{S}^{a_{i}}], \forall j\not=i$. In cases where there are only two groups (group $a_1$ and group $a_2$), there will always be a dominant group. However, when dealing with multiple protected groups, an interesting phenomenon arises where there may not be a dominant group. For instance, if there are three groups (group $a_{1}$, group $a_{2}$ and group $a_{3}$), $a_{1}$ may be dominant compared to $a_{2}$, and $a_{2}$ may be dominant compared to $a_{3}$. However, $a_{3}$ may be dominant compared to $a_{1}$. This lack of transitivity means that $a_{3}$ may be more advantaged than $a_{1}$, even though $a_{1}$ is more advantaged than $a_{2}$ and $a_{2}$ is more advantaged than $a_{3}$. Therefore, there is no single dominant group that is more advantageous than the others. Additional information can be found in Section II in Appendix.
% \begin{proposition}
% Suppose there are three group $a$, $b$ and $c$. Given $\mathrm{Pr}[\mathrm{S}^{a} > \mathrm{S}^{b}] > \mathrm{Pr}[\mathrm{S}^{b} > \mathrm{S}^{a}]$ and $\mathrm{Pr}[\mathrm{S}^{b} > \mathrm{S}^{c}] > \mathrm{Pr}[\mathrm{S}^{c} > \mathrm{S}^{b}]$, it is insufficient for guaranteeing $\mathrm{Pr}[\mathrm{S}^{a} > \mathrm{S}^{c}] > \mathrm{Pr}[\mathrm{S}^{c} > \mathrm{S}^{a}]$.
% \label{prop:multiple}
% \end{proposition}

\subsection{Problem Formulation}
From the above definitions, we can see the utility (measured by AUC as in Eq.(\ref{eq:AUC})) and fairness (e.g., measured by $\Delta$URF in Eq.(\ref{eq:urf})) of the ranking function $R$ are essentially determined by the ordering of data samples induced by the predicted ranking scores. For convenience, we will use xAUC as the fairness metric with two groups in our detailed derivations. The same procedure can be similarly developed for other fairness metrics with multiple protected groups.

Suppose we use $\operatorname{p}^{a}$ and $\operatorname{p}^{b}$ to represent the data sample sequences with their ranking scores ranked in descending order. That is, $\operatorname{p}^{a}=[\operatorname{p}^{a(1)}, \operatorname{p}^{a(2)}, ..., \operatorname{p}^{a(n^a)} ]$ with $R(\operatorname{X}_{\operatorname{p}^{a(i)}},a)\geqslant R(\operatorname{X}_{\operatorname{p}^{a(j)}},a)$ if $0\leqslant i<j\leqslant n^a$, and $\operatorname{p}^{b}$ is defined in the same way, then we have the following definition.
\begin{definition}[Cross-Group Ordering \emph{O}]
Given ordered instance sequences $\operatorname{p}^{a}$ and $\operatorname{p}^{b}$, the \emph{cross-group ordering} $o(\operatorname{p}^{a}, \operatorname{p}^{b})$ defines a ranked list combining the instances in groups $a$ and $b$ while keeping within group instance ranking orders preserved.
\end{definition}
One example of such cross-group ordering is:

$o(\operatorname{p}^{a}, \operatorname{p}^{b})$=$[ \operatorname{p}^{a(1)}, \operatorname{p}^{b(1)}, \operatorname{p}^{a(2)}, ..., \operatorname{p}^{a(n^{a})},..., \operatorname{p}^{b(n^{b})}] $. From the above definitions we can see that we only need cross-group ordering $o(\operatorname{p}^{a}, \operatorname{p}^{b})$ to estimate both algorithm utility measured by AUC and ranking fairness measured by $\Delta$xAUC, i.e., we do not need the actual ranking scores. With this definition, we have the following proposition.

\begin{proposition}
Given ordered instance sequences $\operatorname{p}^{a}$ and $\operatorname{p}^{b}$, for supervised fairness metric (xAUC and PRF) and unsupervised fairenss metric (URF), there exists a crossing-group ordering  $o(\operatorname{p}^{a}, \operatorname{p}^{b})$ that can achieve $\Delta \mathrm{URF} \leq \min(1/n^{a}, 1/n^{b})$, $\Delta \operatorname{xAUC} \leq \min(\max({1}/{n_1^b}, {1}/{n_0^b}), \max({1}/{n_1^a}, {1}/{n_0^a}))$ or $\Delta \operatorname{PRF} \leq \min(\max({n^b_0}/{(n_1^a n_0)}, {1}/{n_0}), \max({n^a_0}/{(n_1^b n_0)}, {1}/{n_0}))$.
\label{pro1}
\end{proposition}

The proof of Proposition~\ref{pro1} is provided in Appendix. From Proposition~\ref{pro1}, if we only care about ranking fairness, we can achieve a relatively low disparity by a trivial method introduced in Appendix. Our proposal in this paper is to look for an optimal cross-group ordering $o^*(\operatorname{p}^{a}, \operatorname{p}^{b})$, which can achieve ranking fairness and maximally maintain algorithm utility, through post-processing.

One important issue to consider is that the cross-group ordering that achieves the same level of ranking fairness is not unique. In Figure~\ref{fig:samples}(a), we demonstrate an illustrative example with 9 samples showing that different cross-group ordering can result in different AUCs with the same $\Delta$xAUC. The middle row in Figure~\ref{fig:samples}(a) shows the original predicted ranking scores and their induced sample ranking, which achieves a ranking disparity $\Delta\mathrm{xAUC}=0.75$. The top row shows one cross-group ordering with ranking disparity $\Delta\mathrm{xAUC} = 0$ and algorithm utility $\mathrm{AUC}=0.56$. The bottom row shows another cross-group ranking with $\Delta\mathrm{xAUC} = 0$ but $\mathrm{AUC}=0.83$.
\begin{figure}[h!]
  \centering
  \includegraphics[width=1.0\columnwidth]{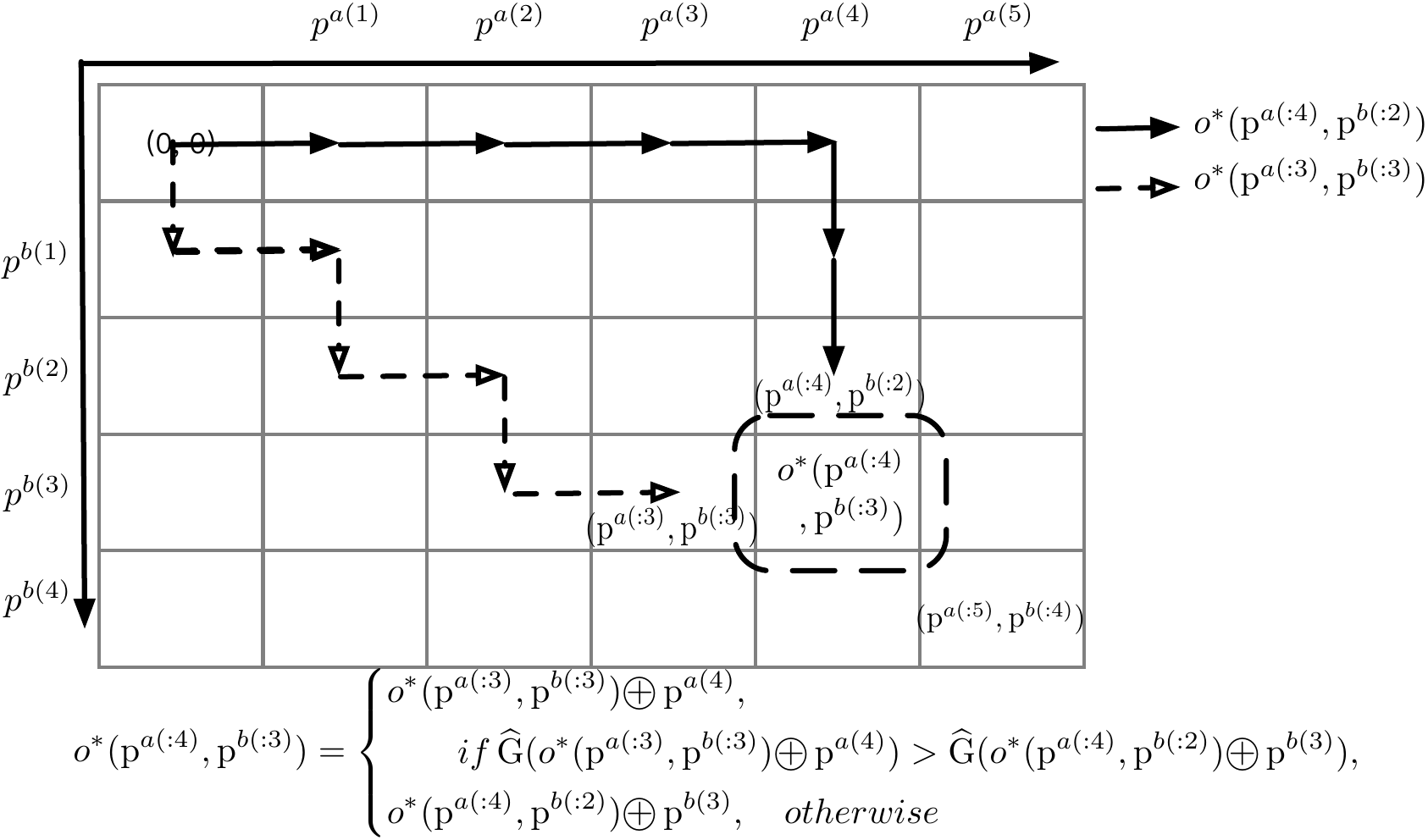}
  \label{fig:optimize_sub_pro}
  \caption{Illustration of how our proposed method optimizes such relative ordering.}
\end{figure}
Our goal is to identify an optimal cross-group ordering $o^*(\operatorname{p}^{a}, \operatorname{p}^{b})$ that leads to the minimum ranking disparity (measured by $\Delta$xAUC) with algorithm utility (measured by AUC) maximally maintained. We can maximize the following objective
\begin{equation}
    \label{eq:optimizationobj}
    \mathrm{J}(o(\operatorname{p}^{a}, \operatorname{p}^{b}))=\mathrm{AUC}^{o}(o(\operatorname{p}^{a}, \operatorname{p}^{b}))- \lambda \cdot\Delta\mathrm{xAUC}^{o}(o(\operatorname{p}^{a}, \operatorname{p}^{b})),
\end{equation}
where we use $\text {AUC}^{o}(o(\operatorname{p}^{a}, \operatorname{p}^{b}))$ to denote the AUC induced by the ordering $o(\operatorname{p}^{a}, \operatorname{p}^{b})$, which is calculated in the same way as in Eq.(\ref{eq:AUC}) if we think of $R(X_i)$ returning the rank of $X_i$ instead of the actual ranking score. Please note that in the rest of this paper we will use similar notations for xAUC without causing further confusion. Similarly, $\Delta\mathrm{xAUC}^{o}(o(\operatorname{p}^{a}, \operatorname{p}^{b}))$ is the ranking disparity induced by ordering $o(\operatorname{p}^{a}, \operatorname{p}^{b})$ calculated as in Eq.(\ref{eq:delta_xAUC}). Then we have the following proposition.
\begin{proposition}
The objective function in Eq.(\ref{eq:optimizationobj}) is equivalent to:
\begin{equation}
\label{eq:G}
\begin{aligned}
&\mathrm{G}(o(\operatorname{p}^{a}, \operatorname{p}^{b}))= \frac{k_{a,b}}{k} \cdot \operatorname{xAUC}^{o}(o(\operatorname{p}^{a}, \operatorname{p}^{b}))\\
&+ \frac{k_{b,a}}{k}  \cdot \operatorname{xAUC}^{o}(o(\operatorname{p}^{b}, \operatorname{p}^{a})) - \lambda\cdot\Delta\mathrm{xAUC}^{o}(o(\operatorname{p}^{a}, \operatorname{p}^{b})),
\end{aligned}
\end{equation}
where $k_{a,b} = n^{a}_{1}n^{b}_{0}$, $k_{b,a} = n^{a}_{0} n^{b}_{1}$, $k = n_{0}n_{1}$, $\operatorname{xAUC}^{o}(o(\operatorname{p}^{a}, \operatorname{p}^{b}))$ indicates $ \operatorname{xAUC}(a,b)$ in Eq.~\ref{eq:xAUC} ($\operatorname{xAUC}^{o}(o(\operatorname{p}^{b}, \operatorname{p}^{a}))$ indicates $\operatorname{xAUC}(b,a)$ in Eq.~\ref{eq:xAUC}), and $\lambda$ is the hyperparameter trading off the algorithm utility and ranking disparity.
\label{prop:3}
\end{proposition}
The proof of Proposition~\ref{prop:3} is provided in Appendix.
\section{Algorithm}
\label{sec:4}
In this section, we will detailly introduce our proposed method \texttt{xOrder} for fair bipartite ranking with two protected groups. Then we will extend \texttt{xOrder} to fair ranking problems with multiple protected groups.
\subsection{xOrder for Fair Ranking with Two Protected Groups}
To intuitively understand the post-processing process, we treat the cross-group ordering as a path from the higher-ranked instances to lower-ranked instances. With the example shown in Figure \ref{fig:samples}(a), we demonstrate the original ordering of those samples induced by their predicted ranking scores (middle row of Figure \ref{fig:samples}(a)) on the top of Figure~\ref{fig:samples}(b), where the direction on the path indicates the ordering. The ordering corresponding to the bottom row of Figure~\ref{fig:samples}(a) is demonstrated at the bottom of Figure~\ref{fig:samples}(b).
\begin{algorithm}
\SetAlgoLined
\caption{xOrder: optimize the cross-group ordering with post-processing}
\label{alg:cross-group ordering}
\KwIn{$\lambda$, the ranking scores $\operatorname{S}^{a}$, $\operatorname{S}^{b}$ from a predictive ranking function $R$}

Sort the ranking scores $\operatorname{S}^{a}$ and $\operatorname{S}^{b}$ in descending order. Get the instances ranking $\operatorname{p}^{a} = [\operatorname{p}^{a(1)},\operatorname{p}^{a(2)},...\operatorname{p}^{a(n^{a})}]$ and $\operatorname{p}^{b} = [\operatorname{p}^{b(1)},\operatorname{p}^{b(1)},...\operatorname{p}^{b(n^{b})}]$\\

Initialize cross-group ordering $o^{*}(\operatorname{p}^{a(:i)}, \operatorname{p}^{b(:0)})$ $(0 \leq i \leq n^{a})$, $o^{*}(\operatorname{p}^{a(:0)}, \operatorname{p}^{b(:j)})$ $(0 \leq j \leq n^{b})$,\\
\ForAll{i = 1, 2, 3... $n^{a}$}{
  \ForAll{j = 1, 2, 3...$n^{b}$}{
  Calculate $\operatorname{\widehat{G}}(o^{*}(\operatorname{p}^{a(:i-1)}, \operatorname{p}^{b(:j)}) \textcircled{+} \operatorname{p}^{a(:i)})$, $\operatorname{\widehat{G}}(o^{*}(\operatorname{p}^{a(:i)}, \operatorname{p}^{b(:j-1)}) \textcircled{+} \operatorname{p}^{b(:j)})$ using the Eq.(\ref{eq:G1})\\
  Update $o^{*}(\operatorname{p}^{a(:i)}, \operatorname{p}^{b(:j)})$ according to the Eq.(\ref{G2})}
}
\KwOut{the learnt cross-group ordering $o^{*}(\operatorname{p}^{a}, \operatorname{p}^{b})$}
\end{algorithm}
With this analogy, the optimal cross-group ordering $o^{*}(\operatorname{p}^{a}, \operatorname{p}^{b})$ can be achieved by a path finding process. The path must start from $\operatorname{p}^{a(1)}$ or $\operatorname{p}^{b(1)}$, and end with $\operatorname{p}^{a(n^a)}$ or $\operatorname{p}^{b(n^b)}$. Each instance in $\operatorname{p}^{a}$ and $\operatorname{p}^{b}$ can only appear once in the final path, and the orders of the instances in the final path must be the same as their orders in $\operatorname{p}^a$ and $\operatorname{p}^b$. The path can be obtained through a dynamic programming process. In particular, we first partition the entire decision space into a $(n^b+1)\times (n^a+1)$
grid. Each location $(i,j)~0\leqslant i\leqslant n^a,0 \leqslant j\leqslant n^b$ on the lattice corresponds to a decision step on determining whether to add $p^{a(i)}$ or $p^{b(j)}$ into the current path, which can be determined with the following rule:
\begin{equation}
\label{G2}
\begin{aligned}
& \text{Given} \quad o^{*}(\operatorname{p}^{a(:i-1)}, \operatorname{p}^{b(:j)}), o^{*}(\operatorname{p}^{a(:i)}, \operatorname{p}^{b(:j-1)})\\
& \text{if:}\ \widehat{\operatorname{G}}(o^*(\operatorname{p}^{a(:i-1)}, \operatorname{p}^{b(:j)})\textcircled{+} \operatorname{p}^{a(i)})>\widehat{\operatorname{G}}(o^*(\operatorname{p}^{a(:i)}, \operatorname{p}^{b(:j-1)})\textcircled{+} \operatorname{p}^{b(j)})\\
& \quad \quad o^{*}(\operatorname{p}^{a(:i)}, \operatorname{p}^{b(:j)}) = o^{*}(\operatorname{p}^{a(:i-1)}, \operatorname{p}^{b(:j)})\textcircled{+} \operatorname{p}^{a(i)};\\
& \text{otherwise:}\ o^{*}(\operatorname{p}^{a(:i)}, \operatorname{p}^{b(:j)}) = o^{*}(\operatorname{p}^{a(:i)}, \operatorname{p}^{b(:j-1)}) \textcircled{+} \operatorname{p}^{b(j)},
\end{aligned}
\end{equation}
where $\operatorname{p}^{a(:i)}$ represents the first $i$ elements in $\operatorname{p}^a$ ($\operatorname{p}^{a(:i-1)}$, $\operatorname{p}^{b(:j-1)}$ and $\operatorname{p}^{b(:j)}$ are similarly defined). $o^{*}(\operatorname{p}^{a(:i-1)}, \operatorname{p}^{b(:j)}) \textcircled{+} \operatorname{p}^{a(i)}$ means appending $\operatorname{p}^{a(i)}$ to the end of $o^*(\operatorname{p}^{a(:i-1)}, \operatorname{p}^{b(:j)})$. The value of function $\widehat{\mathrm{G}}(o(\operatorname{p}^{a(:i)},\operatorname{p}^{b(:j)}))$ is defined as follows
\begin{equation}
\label{eq:G1}
\begin{aligned}
&\widehat{\operatorname{G}}(o(\operatorname{p}^{a(:i)}, \operatorname{p}^{b(:j)})) = \\
&\frac{k_{a,b}}{k} \cdot {\operatorname{xAUC}}(o(\operatorname{p}^{a(:i)}, \operatorname{p}^{b(:j)}) \textcircled{+} \operatorname{p}^{b(j+1:n^b)})\\
&+\frac{k_{b,a}}{k} \cdot {\operatorname{xAUC}}(o(\operatorname{p}^{b(:j)}, \operatorname{p}^{a(:i)}) \textcircled{+} \operatorname{p}^{a(i+1:n^a)})\\
&-\lambda \cdot |{\operatorname{xAUC}}(o(\operatorname{p}^{a(:i)}, \operatorname{p}^{b(:j)}) \textcircled{+} \operatorname{p}^{b(j+1:n^b)})\\
&-{\operatorname{xAUC}}(o(\operatorname{p}^{b(:j)}, \operatorname{p}^{a(:i)}) \textcircled{+} \operatorname{p}^{a(i+1:n^a)})|,
\end{aligned}
\end{equation}

in which $\widehat{\operatorname{G}}(o(\operatorname{p}^{a(:i)}, \operatorname{p}^{b(:j)})) = \operatorname{G}(o(\operatorname{p}^{a}, \operatorname{p}^{b}))$ in Eq.(\ref{eq:optimizationobj}) when $i = n^a,j=n^b$.

Figure \ref{fig:samples}(b) demonstrates the case of applying our rule to step $i=4,j=3$ in the example shown in Figure \ref{fig:samples}(a).

Algorithm \ref{alg:cross-group ordering} summarized the whole pipeline of identifying the optimal path. In particular, our algorithm calculates a cost for every point $(i,j)$ in the decision lattice as the value of the $\operatorname{\widehat{G}}$ function evaluated on the path reaching $(i,j)$ from $(0,0)$. Please note that the first row ($j=0$) only involves the instances in $a$, therefore the path reaching the points in this row is uniquely defined considering the within-group instance order should be preserved in the path. The decision points in the first column ($i=0$) enjoy similar characteristics. After the cost values for the decision points in the first row and column are calculated, the cost values on the rest of the decision points in the lattice can be calculated iteratively until $i=n^a$ and $j=n^b$.

Algorithm 1 can also be viewed as a process to maximize the objective function in Eq.(\ref{eq:optimizationobj}). It has $O(N^2)$ time complexity as it is a 2-D dynamic programming process. Different $\lambda$ values trade off the algorithm utility and ranking fairness differently and the solution Algorithm 1 converges to is a local optimum. Moreover, we have the following proposition.

\begin{theorem}
xOrder can achieve the global optimal solution of maximizing Eq.(\ref{eq:optimizationobj}) with $\lambda=0$. xOrder has the upper bounds of fairness disparities defined in Eq.(\ref{eq:delta_xAUC}) and Eq.(\ref{eq:deltaPRF}) as $\lambda$ approaches infinity:

\begin{equation}
\begin{aligned}
&\Delta \mathrm{URF} \leq \max \left(\frac{1}{n^{a}}, \frac{1}{n^{b}}\right)\\
&\Delta \mathrm{x} \mathrm{AUC} \leq \max \left(\frac{1}{n_{1}^{a}}, \frac{1}{n_{1}^{b}}\right)\\
&\Delta \mathrm{PRF} \leq \max \left(\frac{n_{0}^{b}}{n_{0} \cdot n_{1}^{a}}, \frac{n_{0}^{a}}{n_{0} \cdot n_{1}^{b}}\right)
\end{aligned}
\end{equation}
\end{theorem}
\revise{The proof of this theorem is provided in Appendix. For a non-trivial trade-off between fairness and utility with a normal $0 < \lambda < + \infty$, if we only care about the disparity of the group $a$ to the group $b$, i.e., the disparity is $\mathrm{xAUC}^{o}(o(p^{a(:i)}, p^{b(:j)} )) - \mathrm{xAUC}^{o}(o(p^{b(:j)}, p^{a(:i)}))$ rather than $\| \mathrm{xAUC}^{o}(o(p^{a(:i)}, p^{b(:j)} )) - \mathrm{xAUC}^{o}(o(p^{b(:j)}, p^{a(:i)})) \|$, the objective is re-formulated as follows
\begin{equation}
\begin{aligned}
\label{eq:optimizationobj_revise}
\mathrm{J}(o(\operatorname{p}^{a}, \operatorname{p}^{b})) &=\mathrm{AUC}^{o}(o(\operatorname{p}^{a}, \operatorname{p}^{b})) \\
& - \lambda (\mathrm{xAUC}^{o}(o(p^{a}, p^{b} )) - \mathrm{xAUC}^{o}(o(p^{b}, p^{a}))).
\end{aligned}
\end{equation}
\texttt{xOrder} could also achieve the global optimum for the objective in Eq. (\ref{eq:optimizationobj_revise}).}

\noindent\textbf{The testing phase}. Since the learned ordering in the training phase cannot be directly used in the testing stage, we propose to transfer the information of the learned ordering by rearranging the ranking scores of the disadvantaged group (which is assumed to be group $b$ without the loss of generality). If we assume the distribution of predicted ranking scores is the same on the training and test sets, the same scores will have the same quantiles in both sets. It means that by rearranging the ranking scores through interpolation, we can transfer the ordering learned by \texttt{xOrder} on the training set. In particular, the process contains two steps:
\begin{enumerate}[leftmargin=*]
    \item \underline{\emph{Rank score adjustment for the group b in training data}}. Fixing the ranking scores for training instances in the group $a$, the adjusted ranking scores for training instances in group $b$ will be obtained by uniform linear interpolation according to their relative positions in the learned ordering. For example, if we have an ordered sequence $(\operatorname{p}^{a(1)},\operatorname{p}^{b(1)},\operatorname{p}^{b(2)},\operatorname{p}^{a(2)})$ with the ranking scores for $\operatorname{p}^{a(1)}$ and $\operatorname{p}^{a(2)}$ being 0.8 and 0.5, then the adjusted ranking scores for $\operatorname{p}^{b(1)}$ and $\operatorname{p}^{b(2)}$ being 0.7 and 0.6.

    \item \underline{\emph{Rank score adjustment for the group b in test data}}. For testing instances, we follow the same practice of just adjusting the ranking scores of the instances from group $b$ and keeping the ranking scores for instances from the group $a$ unchanged. \sen{We propose a proportional interpolation for the adjustment process which has O(N) time complexity. In particular, we first rank training instances from group $b$ according to their raw unadjusted ranking scores to get an ordered list. Then the adjusted ranking scores for testing instances in $b$ can be obtained by a linear transformation. For example, if we want to adjust the testing original score $\operatorname{p_{te}}^{b(i)}$ given original training ordered sequence $(\operatorname{p}^{b(1)},\operatorname{p}^{b(2)}$ being 0.8 and 0.5 and the training adjusted ordered sequence $(\operatorname{\hat{p}}^{b(1)},\operatorname{\hat{p}}^{b(2)}$ being 0.7 and 0.4, then the quantile of the testing adjusted ranking score $\operatorname{\hat{p}_{te}}^{b(i)}$ in [0.4, 0.7] equals to the quantile of $\operatorname{p_{te}}^{b(i)}$ in [0.5, 0.8]:}
    $\frac{0.7 - \operatorname{\hat{p}_{te}}^{b(i)}}{0.7-0.4} = \frac{0.8 - \operatorname{p_{te}}^{b(i)}}{0.8-0.5}.$
\end{enumerate}
From the above process, we transfer the learned ordering from the training set to the test set by interpolation. We also try other operations to achieve interpolation (e.g. ranking uniform linear interpolation) and get similar experimental results.

\subsection{Fair Bipartite Ranking with Multiple Protected Groups}
\label{sec:3-groups}
\begin{figure}[h]
    \centering
    \includegraphics[width=0.95\columnwidth]{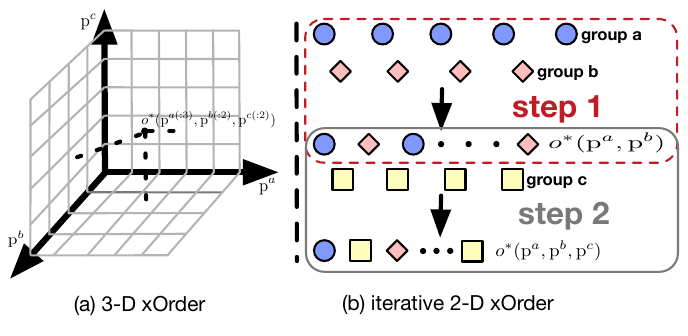}
    \caption{(a) Illustration of 3-D \texttt{xOrder}; (b) An iterative \texttt{xOrder} for fair ranking with three protected groups. Step 1, \texttt{xOrder} determines the cross-group ordering $o^{*}(\mathrm{p}^{a}, \mathrm{p}^{b})$; Step 2, \texttt{xOrder} determines the final optimal crossing-group ordering $o^{*}(\mathrm{p}^{a}, \mathrm{p}^{b}, \mathrm{p}^{c})$.}
    \label{fig:3-D-xOrder}
\end{figure}
To achieve a fair ranking with multiple protected groups, we propose two variants of \texttt{xOrder}, \textbf{K-D xOrder} and \textbf{iterative xOrder}. Recall that when there are two protected groups, the optimization procedure of \texttt{xOrder} is a 2-D dynamic programming process as shown in Figure 3. In fact, \texttt{xOrder} could be naturally extended to achieve a fair ranking with K protected groups, which behaves as a K-D dynamic programming.

\noindent \textbf{K-D xOrder.} Suppose there are three protected groups $a$, $b$ and $c$ with the ordered instance sequences $\mathrm{p}^{a}$, $\mathrm{p}^{b}$ and $\mathrm{p}^{c}$, the cross-group ordering is denoted as $o(\mathrm{p}^{a}, \mathrm{p}^{b}, \mathrm{p}^{c})$, the disparity measured by xAUC metric is formulated as follows
\begin{equation}
\begin{aligned}
& \Delta \mathrm{xAUC}^{o}(o(\mathrm{p}^{a}, \mathrm{p}^{b}, \mathrm{p}^{c})) = \\
& \max\left\{\Delta \mathrm{xAUC}^{o}(o(\mathrm{p}^{{i}}, \mathrm{p}^{j})) \right\}, \ i, j \in \left\{a,b,c \right\} \ and \ i\not=j.
\end{aligned}
\label{eq:xauc_abc}
\end{equation}
And the objective to be maximized is formulated as follows
\begin{equation}
\begin{aligned}
& \mathrm{J}(o(\mathrm{p}^{a}, \mathrm{p}^{b}, \mathrm{p}^{c})) = \\
& \mathrm{AUC}^{o}(o(\mathrm{p}^{a}, \mathrm{p}^{b}, \mathrm{p}^{c})) - \lambda \cdot \Delta \mathrm{xAUC}^{o}(o(\mathrm{p}^{a}, \mathrm{p}^{b}, \mathrm{p}^{c})).
\end{aligned}
\label{eq:xauc_abc_obj}
\end{equation}
The optimization of the cross-group ordering among the three groups is illustrated in Figure~\ref{fig:3-D-xOrder}(a).

From Figure~\ref{fig:3-D-xOrder}(a), similar to the optimization process with two groups shown in Figure 3, to update the optimal cross-group ordering $o^{*}(\mathrm{p}^{a(:i)}, \mathrm{p}^{b(:j)}, \mathrm{p}^{c(:k)})$, we maximize the value function $\widehat{\mathrm{G}}\left(o\left(\mathrm{p}^{a(: i)}, \mathrm{p}^{b(: j)}, \mathrm{p}^{c(: k)}\right)\right)$ by determining whether to add $a(i)$, $b(j)$ or $c(k)$ into the current path.

\noindent \textbf{iterative xOrder.} As a 3-D dynamic programming process, it has O$(N^{3})$ time complexity (N denotes the number of samples in each group). To improve the efficiency, we propose an iterative \texttt{xOrder} to balance the fairness and utility, which has O$(N^{2})$ time complexity with as many groups.

We propose to reach an optimal ordering $o^{*}(\mathrm{p}^{a}, \mathrm{p}^{b}, \mathrm{p}^{c})$ for Eq.(\ref{eq:xauc_abc_obj}) by optimizing the following objectives
\begin{equation}
\begin{aligned}
& step \ 1: \quad \mathrm{J}(o(\mathrm{p}^{a}, \mathrm{p}^{b})) = \\
& \mathrm{AUC}^{o}(o(\mathrm{p}^{a}, \mathrm{p}^{b})) - \lambda \cdot \Delta \mathrm{xAUC}^{o}(o(\mathrm{p}^{a}, \mathrm{p}^{b})) \\
& step \ 2: \quad \mathrm{J}(o( o^{*}(\mathrm{p}^{a}, \mathrm{p}^{b}), \mathrm{p}^{c})) = \\
& \mathrm{AUC}^{o}(o(o^{*}(\mathrm{p}^{a}, \mathrm{p}^{b}), \mathrm{p}^{c})) - \lambda \cdot \Delta \mathrm{xAUC}^{o}(o(o^{*}(\mathrm{p}^{a}, \mathrm{p}^{b}), \mathrm{p}^{c})).
\end{aligned}
\label{eq:xauc_abc_obj_2}
\end{equation}

As shown in Figure~\ref{fig:3-D-xOrder}(b), for the three groups $a$, $b$ and $c$, we first obtain $o^{*}(\mathrm{p}^{a}, \mathrm{p}^{b})$ by 2-D \texttt{xOrder}, which is the \emph{step 1} in Eq.(\ref{eq:xauc_abc_obj_2}). Then we optimize $o(o^{*}(\mathrm{p}^{a}, \mathrm{p}^{b}), \mathrm{p}^{c})$ to maximize $\mathrm{J}(o( o^{*}(\mathrm{p}^{a}, \mathrm{p}^{b}), \mathrm{p}^{c}))$ by 2-D \texttt{xOrder} to finalize the ordering of the three groups.

By an iterative operation, we significantly reduce the time complexity to $O(K^{2}N^{2})$ for $K$ protected groups with $N$ samples in each group \footnote{when the sample size varies in different groups, the time complexity is less than $O(K^{2}N_{max}^{2})$ and $N_{max}$ denotes the largest sample size of all groups.} Experimental results in Section~\ref{sec:5} demonstrate that iterative \texttt{xOrder} maintains the best trade-off between fairness and utility compared with baselines, and has a practical time complexity and memory requirement. More information could be found in Appendix.

\section{Experiment}
\label{sec:5}
\subsection{Data Sets}
\noindent\textbf{Data sets}. We conduct experiments on 4 popular benchmark data sets and two real-world clinical data sets for studying algorithm fairness. For each data set, we randomly select 70\% of the data set as the train set and the remaining as the test set following the setting in~\cite{kallus2019fairness}. The basic information for these data sets is summarized in Table \ref{table:summary_eicu}, where $n$ and $p$ are the numbers of instances and features, and CHD is the abbreviation of coronary heart disease. To study ranking fairness in real-world clinical ranking prediction scenarios, we analyze the unfair phenomena and conduct experiments on two clinical data sets MIMIC-III and eICU. As in Table \ref{table:summary_eicu}, MIMIC-III a real-world electronic health record repository for ICU patients~\cite{johnson2016mimic}. The data set was preprocessed as in~\cite{harutyunyan2019multitask} with $n$ = 21,139 and $p$ = 714. Each instance is a specific ICU stay. eICU is another real-world dataset of ICU electronic health records~\cite{pollard2018eicu}. It is was preprocessed with $n$ = 17,402 and $p$ = 60. We consider the same setting of the label and protected variable for MIMIC and eICU. For label $\operatorname{Y}$, we consider in-hospital mortality and prolonged length of stay (whether the ICU stay is longer than 1 week). For protected variable $\operatorname{A}$, we consider gender (male, female) and ethnicity (white, non-white). For the setting of multi-group ranking fairness, we use age (age$\leq$25, 25$\leq$age$\leq$40 and 40$\leq$age) as the sensitive attribute following the work in ~\cite{kamiran2009classifying,friedler2019comparative,zemel2013learning,jiang2020wasserstein} and conduct experiments on Adult and COMPAS datasets.
\begin{table}
\tiny
    \caption{Summary of benchmark data sets}.
    \label{table:summary_eicu}
    \centering
    \begin{tabular}{c|c|c|c|c}
        \hline
        Dataset & $n$ & $p$ & $\mathrm{A}$ & $\mathrm{Y}$ \\
        \hline
        COMPAS\cite{angwin2016machine} & 6,167 & 400 & Race(white, non-white) & Non-recidivism within 2 years \\
        Adult\cite{kohavi1996scaling} & 30,162 & 98 & Race(white, non-white) & Income $\geq$ 50K\\
        Framingham~\cite{levy199950} & 4,658 & 7 & Gender(male,female) & 10-year CHD incidence\\
        MIMIC\cite{johnson2016mimic} & 21,139 & 714 & Gender(male,female) & Mortality\\
        MIMIC\cite{johnson2016mimic} & 21,139 & 714 & Race(white, non-white) & Prolonged length of stay\\
        eICU\cite{pollard2018eicu} & 17,402 & 60 & Race(white, non-white) & Prolonged length of stay\\
        \hline
    \end{tabular}
\vspace{-.5em}
\end{table}
\subsection{Baselines}
In this paper,$\Delta$URF, $\Delta \mathrm{xAUC}$ and $\Delta \mathrm{PRF}$ are used as ranking fairness metrics and $\mathrm{AUC}$ is used to measure the algorithm utility. To verify the validity and applicability of our algorithm, we adopt 2 base models: linear model \cite{narasimhan2020pairwise} \cite{kallus2019fairness} and rankboost~\cite{freund2003efficient}. The algorithms that originally proposed xAUC~\cite{kallus2019fairness} PRF~\cite{beutel2019fairness} as well as the framework proposed in ~\cite{narasimhan2020pairwise} are evaluated as baselines, which are denoted as \textbf{post-logit}~\cite{kallus2019fairness}, \textbf{corr-reg}~\cite{beutel2019fairness} and \textbf{opti-tfco}~\cite{narasimhan2020pairwise}. We also report the performance obtained by the two original base models without fairness considerations (called \textbf{unadjusted}). post-logit and \texttt{xOrder} are post-processing algorithms and can be used to all base models.

We propose the following training procedure for the linear model. For corr-reg, the model is trained by optimizing the weighted sum of the cross entropy loss and fairness regularization originally proposed in ~\cite{beutel2019fairness} with gradient descent. Since the constrained optimization problem in opti-tfco can not be solved with gradient descent, we optimize it with a specific optimization framework called tfco \cite{cotter2019two} as the original implementation did \cite{narasimhan2020pairwise}. For post-processing algorithms including post-logit and \texttt{xOrder}, we train the linear model on training data without any specific considerations on ranking fairness to obtain the unadjusted prediction ranking scores. The model is optimized with two optimization methods: 1)optimizing cross entropy with gradient descent; 2) solving an unconstrained problem to maximize utility with tfco. For the rankboost, as it is optimized by boosting iterations, it is not compatible with corr-reg and opti-tfco. We only report the results of post-logit and \texttt{xOrder}.

Note that opti-tfco is originally proposed to address the supervised fairness metric, which may not be directly used for $\mathrm{URF}$ metric. The total $\mathrm{AUC}$ and ranking fairness metrics on test data are reported. In addition, we plot a curve showing the trade-off between ranking fairness and algorithm utility for \texttt{xOrder}, corr-reg and opti-tfco with the varying trade-off parameter. We repeat each setting of the experiment ten times and report the average results. To make the comparisons fair, we report the results of different methods under the same base model separately.
\begin{figure*}[htbp]
\setlength{\abovecaptionskip}{0pt}
\setlength{\belowcaptionskip}{-0.0cm}
    \centering
    \includegraphics[width=1.9\columnwidth]{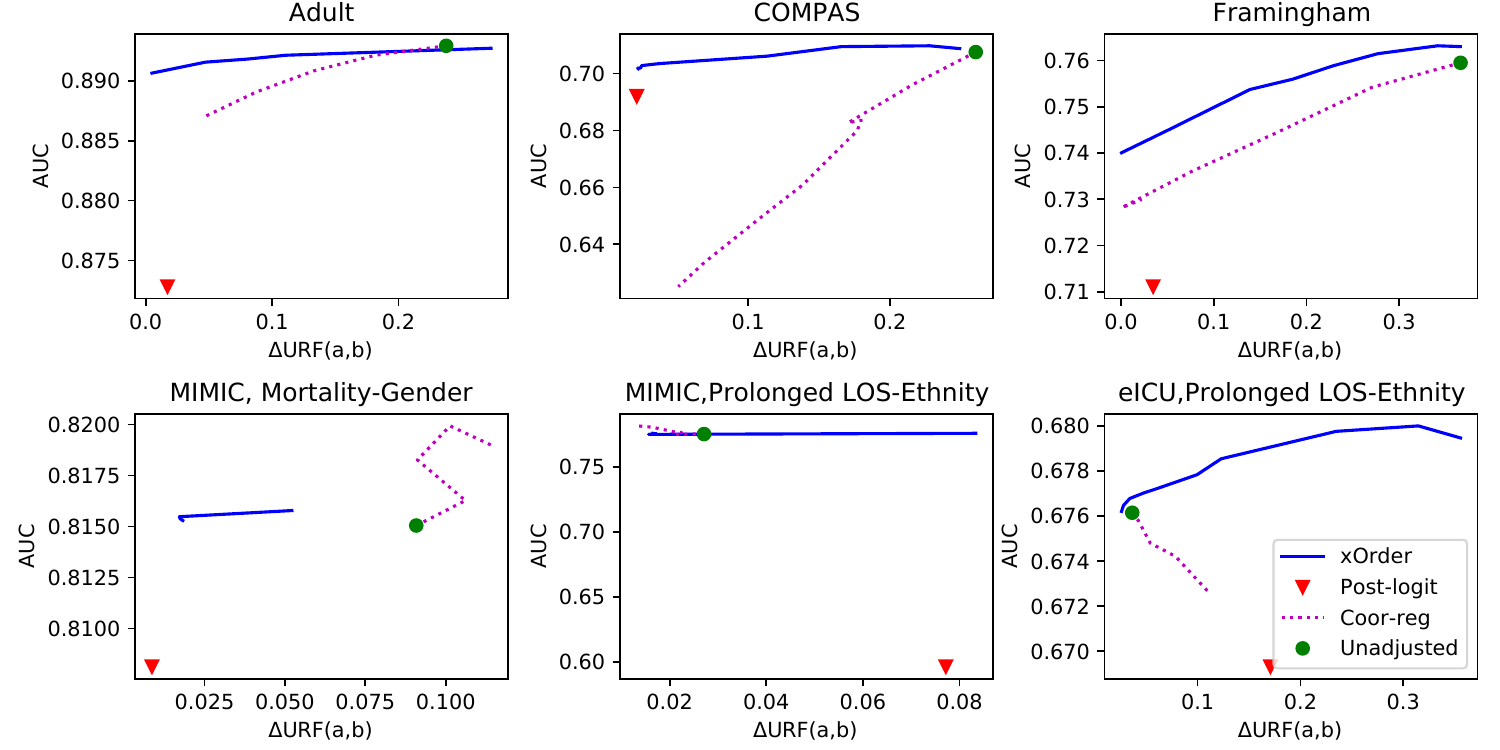}
    \caption{$\mathrm{AUC}$-$\mathrm{\Delta URF}$ trade-off with linear model.}
    \label{fig:lr_urf_result}
\end{figure*}
\begin{figure*}[htbp]
\setlength{\abovecaptionskip}{0pt}
\setlength{\belowcaptionskip}{-0.0cm}
    \centering
    \includegraphics[width=1.95\columnwidth]{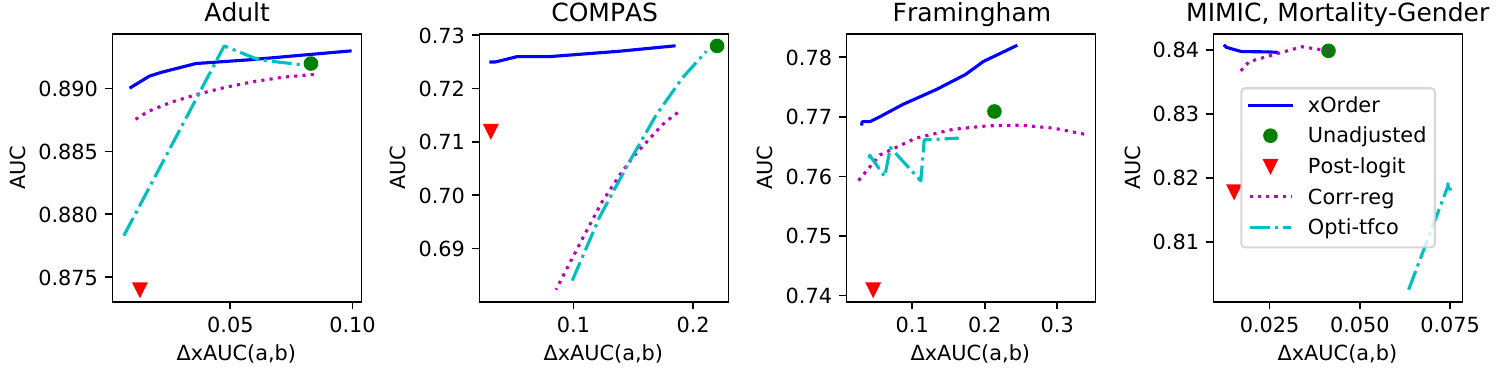}
    \caption{$\mathrm{AUC}$-$\mathrm{\Delta xAUC}$ trade-off with linear model.}
    \label{fig:lr_xauc_result}
\end{figure*}
\begin{figure*}[htbp]
\setlength{\abovecaptionskip}{0pt}
\setlength{\belowcaptionskip}{-.1cm}
    \centering
    \includegraphics[width=1.95\columnwidth]{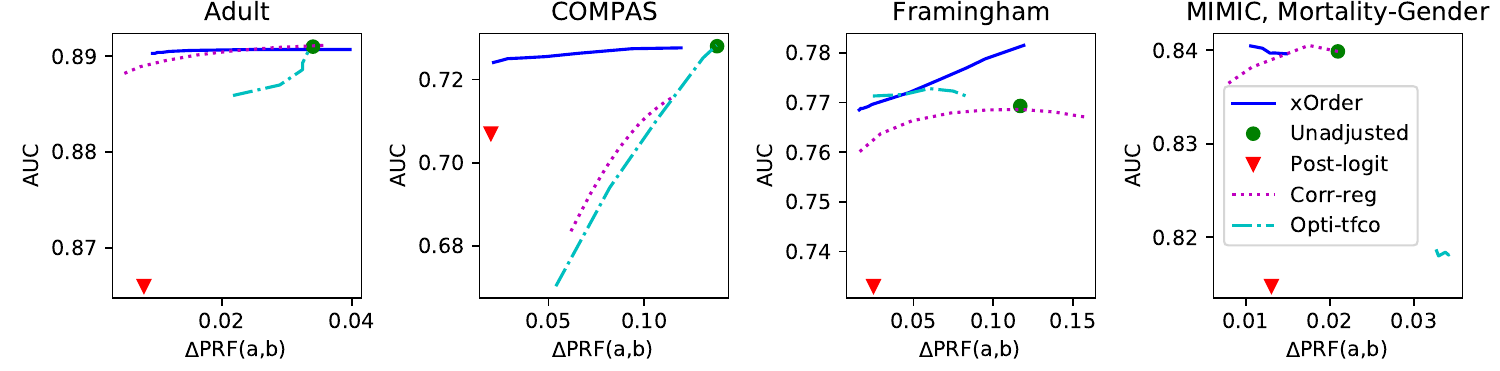}
    \caption{$\mathrm{AUC}$-$\Delta \mathrm{PRF}$ trade-off with linear model.}
    \label{fig:pr_result}
\vspace{-.1em}
\end{figure*}
\begin{figure*}[htbp]
\setlength{\abovecaptionskip}{0pt}
\setlength{\belowcaptionskip}{-0.2cm}
    \centering
    \includegraphics[width=1.95\columnwidth]{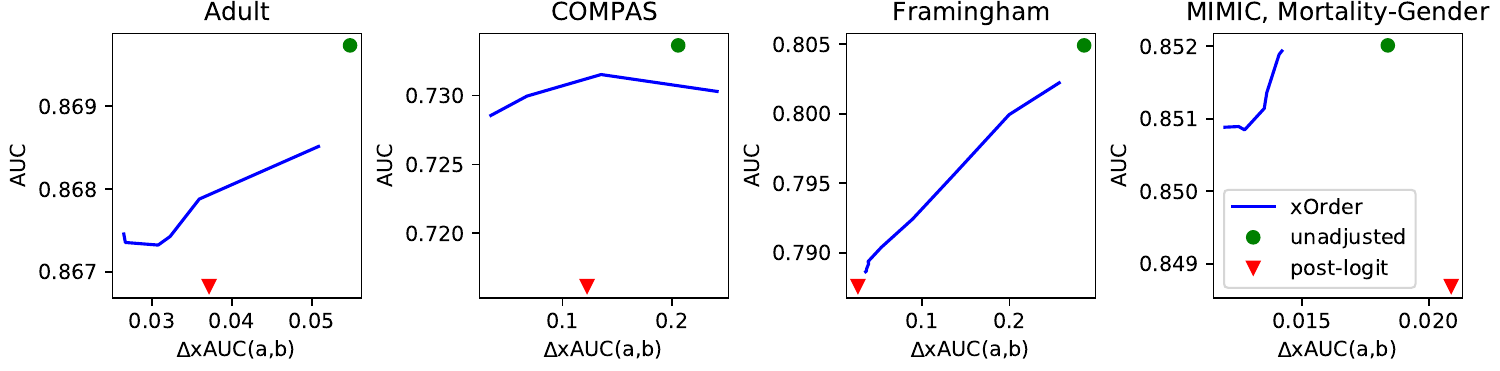}
    \caption{$\mathrm{AUC}$-$\Delta \mathrm{xAUC}$ trade-off with bipartite rankboost model.}
    \label{fig:rb_result}
\vspace{-.2em}
\end{figure*}
\subsection{Evaluation Results}
For the three benchmark data sets (Adult, COMPAS, Framingham), we conduct the experiments with $\operatorname{Y}-\operatorname{A}$ combinations shown in Table \ref{table:summary_eicu}. For MIMIC-III, we show the results as $\operatorname{Y}-\operatorname{A}$ combinations (mortality-gender and prolonged length of stay (LOS)-ethnicity). For eICU, we show the results as $\operatorname{Y}-\operatorname{A}$ combination prolonged length of stay (LOS)-ethnicity. In the experiments, we find that there is no significant unfairness when we focus on prolonged length of stay prediction with sensitive attribute gender. However, we observe an obvious disparity on prolonged length of stay prediction with sensitive attribute ethnicity. We guess this phenomenon is in line with common sense, since the difference in economic conditions is more reflected in race rather than gender for patients living in the ICU.

The results are shown in Figure \ref{fig:lr_urf_result}-\ref{fig:rb_result}. For expression simplicity, we show the result on all data sets using a linear model with the metric $\Delta \mathrm{URF}$ while we show other experiments on 4 data sets. Complete results on all data sets can be found in the supplemental material. In summary, our algorithm achieves superior or comparable performances on all data sets. For the linear model trained with each optimization method, \texttt{xOrder} outperforms all baselines and the complete results with both optimization methods are shown in the supplemental material. For the results of post-processing algorithms with the linear model under two optimization methods, we report them with higher average AUC of the unadjusted results in our main text. It also indicates that post-processing algorithms are model agnostic which has the advantage to choose a better optimization method.
\subsubsection{Linear Model with Unsupervised Metric $\Delta$URF}
We present the experimental results of three methods (\texttt{xOrder}, corr-reg and post-logit) on all datasets in Figure~\ref{fig:lr_urf_result}, in which the base model has a significant disparity on Adult, COMPAS and Framingham datasets. For ranking fairness, \texttt{xOrder} achieves a low disparity on all datasets, while corr-reg and post-logit fail on MIMIC. As a post-processing method, one advantage of \texttt{xOrder} over post-logit is that \texttt{xOrder} can balance the utility and fairness metric within a wide range of $\Delta$URF. For the algorithm utility, \texttt{xOrder} maintains a better trade-off between fairness and utility compared with corr-reg. By adjusting the ordering directly, \texttt{xOrder} optimizes the ranking more flexibly from the results.
\subsubsection{Linear Model with Supervised Metric $\Delta$xAUC and $\Delta$PRF}
The results of linear model with $\Delta$xAUC are shown in Figure~\ref{fig:lr_xauc_result}. All four methods considering ranking fairness (\texttt{xOrder}, corr-reg, post-logit and opti-tfco) are able to obtain lower $\Delta \mathrm{xAUC}$ compared to the unadjusted results. \texttt{xOrder} and post-logit can achieve $\Delta \mathrm{xAUC}$ close to zero on almost all the datasets. This supports proposition 1 empirically that post-processing by changing cross-group ordering has the potential to achieve $\Delta \mathrm{xAUC}$ close to zero. corr-reg and opti-tfco fail to obtain results with low ranking disparities on COMPAS and MIMIC-III. One possible reason is that the correlation regularizer is only an approximation of ranking disparity.

Figure~\ref{fig:pr_result} illustrates the results using a linear model with $\Delta \mathrm{PRF}$ metric on four data sets. The findings are similar to those in Figure \ref{fig:lr_xauc_result}, since $\Delta \mathrm{PRF}$ and $\Delta \mathrm{xAUC}$ are highly correlated on these data sets. \texttt{xOrder} maintains its superiority over all baselines.
\begin{figure}[h]
    \centering
        \includegraphics[width=1.0\columnwidth]{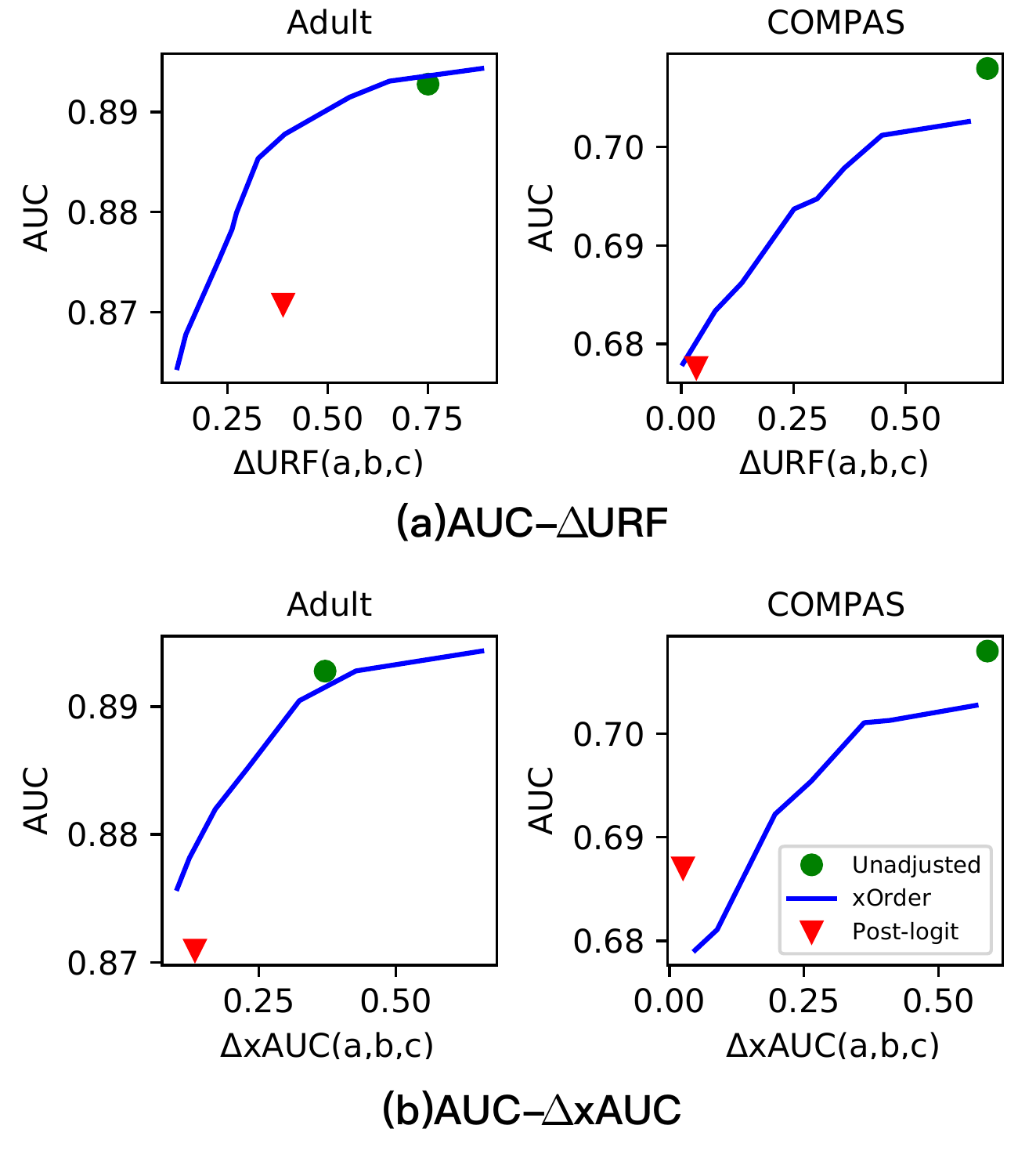}
    \caption{Experimental results on Adult and COMPAS datasets with multiple protected groups. (a).$\mathrm{AUC}$-$\Delta \mathrm{URF}$ trade-off with linear model; (b).$\mathrm{AUC}$-$\Delta \mathrm{xAUC}$ trade-off with linear model.}
    \label{fig:multiple_groups}
\end{figure}
\begin{figure}[h]
    \centering
        \includegraphics[width=1.0\columnwidth]{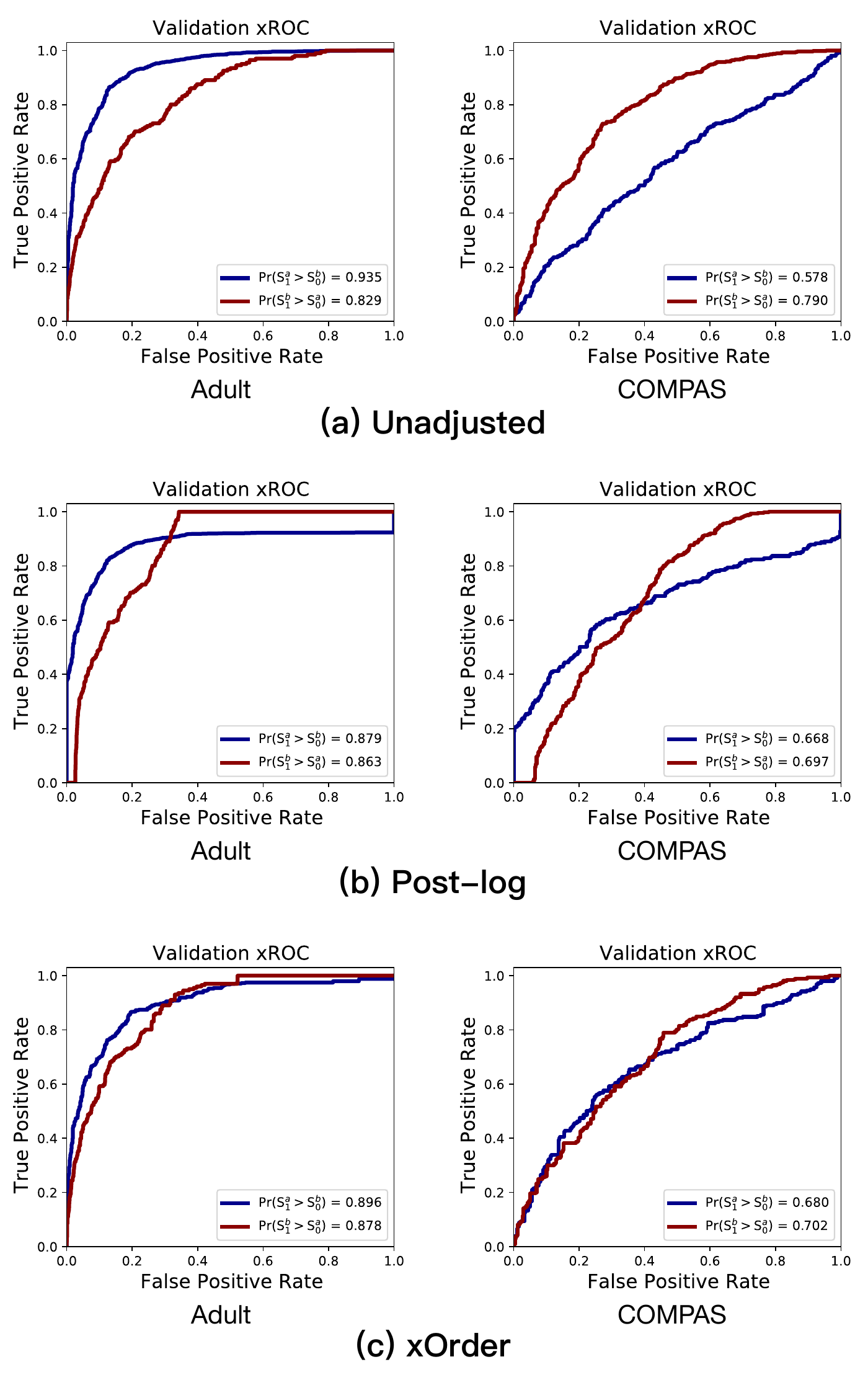}
    \caption{The xROC curves on Adult and COMPAS using a linear model. (a) the risk scores from the base model; (b the risk scores from Post-log; (c) the risk scores from \texttt{xOrder}.}
    \label{fig:xauc_vis}
\end{figure}
Another observation is that \texttt{xOrder} can achieve a better trade-off between utility and fairness. It can obtain competitive or better $\mathrm{AUC}$ under the same level of $\Delta \mathrm{xAUC}$ or $\Delta$URF compared with other methods, especially in the regiment with low disparity. corr-reg performs worse than other methods in COMPAS, while post-logit performs worse in Adult and Fragmingham and opti-tfco performs worse in COMPAS and MIMIC-III (Mortality-Gender). \texttt{xOrder} performs well consistently on all data sets.
\subsubsection{Rankboost with Metric $\Delta \mathrm{xAUC}$}
The results on the bipartite rankboost model and the associated $\Delta \mathrm{xAUC}$ are shown in Figure \ref{fig:rb_result}. It can still be observed that \texttt{xOrder} achieves higher $\mathrm{AUC}$ than post-logit under the same $\Delta \mathrm{xAUC}$. Moreover, post-logit cannot achieve $\Delta \mathrm{xAUC}$ as low as \texttt{xOrder} on adult and COMPAS. This is because the distributions of the prediction scores from the bipartite rankboost model on training and test data are significantly different. Therefore, the function in post-logit which achieves equal $\mathrm{xAUC}$ on training data may not generalize well on test data. Our method is robust against such differences. We empirically illustrate the relation between the distribution of the prediction ranking scores and the generalization ability of post-processing algorithms in the next section.
\subsection{Fair Ranking with Multiple Protected Groups}
We also evaluate the performance of the methods on Adult and COMPAS datasets with three groups (age$\leq$25, 25$\leq$age$\leq$40 and 40$\leq$age. Since corr-reg cannot handle the ranking fairness with multiple groups as well as opti-tfco because of unaccepted time complexity, we display the results of \texttt{xOrder} and post-logit in Figure~\ref{fig:multiple_groups}.

As the disparity of multiple groups is defined as the maximum of disparity of each pair of groups, the base model without considering unfairness could have a severe disparity (e.g., $\Delta \mathrm{URF} = 0.75$ on Adult). Compared with post-logit, \texttt{xOrder} achieves a comparable or superior ranking fairness and algorithm utility from Figure~\ref{fig:multiple_groups}.

\revise{\subsection{More Samples and More Protected Groups}
To validate the practicability of \texttt{xOrder} in the scenario where there are many samples with more protected groups, we conduct a series of experiments on a larger real-world dataset larger eICU~\cite{pollard2018eicu}, which has $n = 127564$ samples with $p = 78$ features. We use race and age as sensitive attributes. For the learning tasks, we consider in-hospital mortality and prolonged length of stay (whether the ICU stay is longer than three days) predictions. We compare the performance with the baselines post-logit and corr-reg.

\begin{figure}[h]
    \centering
    \includegraphics[width=0.95\columnwidth]{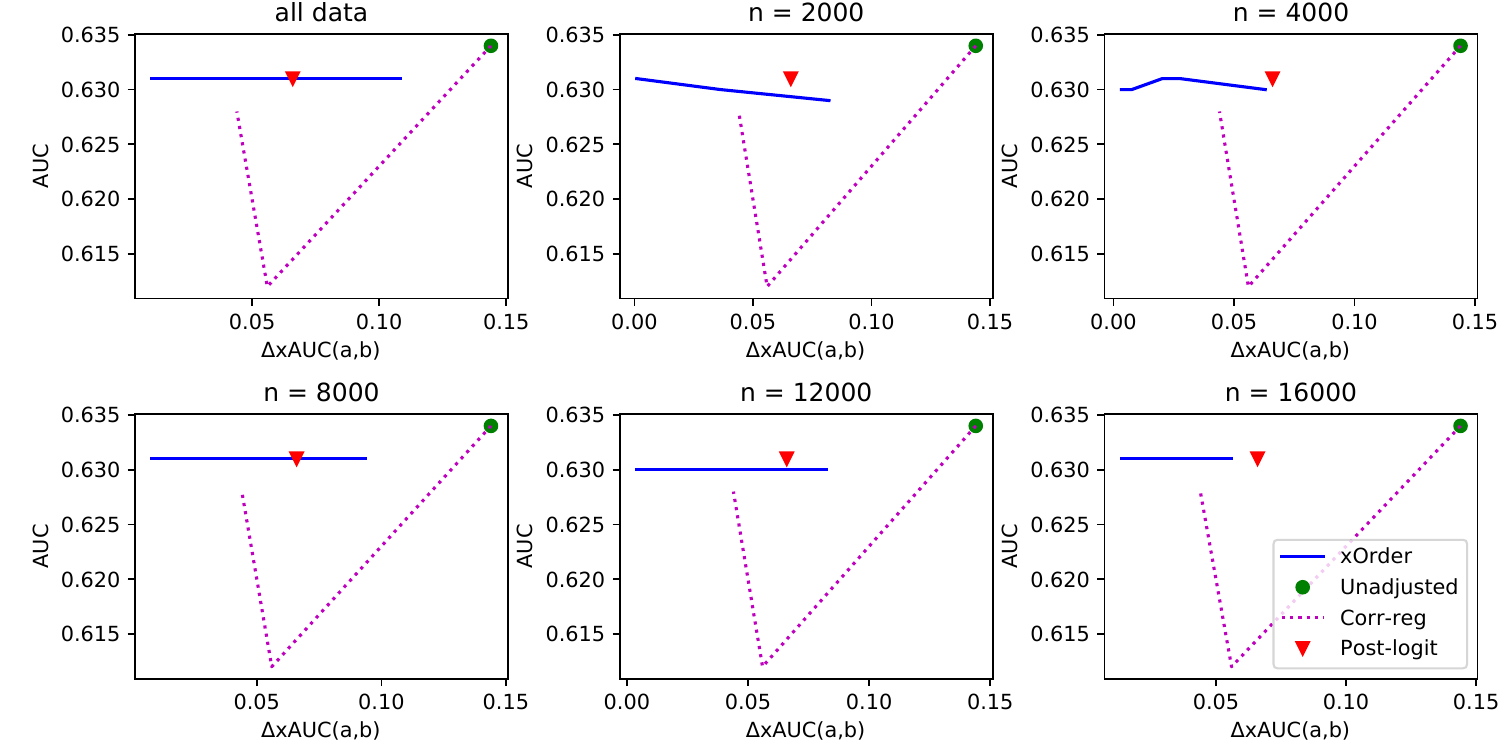}
    \caption{Experimental results on large eICU with the different number of sample sizes (predicting the prolonged length of stay and the sensitive attribute is race).}
    \label{fig:diff_samples}
\end{figure}

Compared with post-logit and corr-reg, \texttt{xOrder} has a higher time complexity (see Section VI in Appendix). Following the advice from the reviewers, we propose to learn an adjustment using \texttt{xOrder} with a subset of the dataset. In particular, we sample data from the ordered ranking with equidistance. For example, given an ordering arranged from large predicted scores to small predicted scores with 1000 samples, e.g., [$a(1)$, $a(2)$, $a(3)$, ..., $a(999)$, $a(1000)$], we sample 100 samples with equidistance [$a(1)$, $a(11)$, $a(21)$, ..., $a(981)$, $a(991)$]. We conduct experiments to predict the prolonged length of stay on large eICU with two protected groups. The baselines use all data samples. The experimental results are shown in Figure~\ref{fig:diff_samples}.

From Figure~\ref{fig:diff_samples}, \texttt{xOrder} maintains the best trade-off between fairness and utility compared with baselines. When there are 2000 samples, \texttt{xOrder} learns an effective adjustment with a small subset, which achieves a comparable performance compared with the performance of \texttt{xOrder} using all data samples. Experimental results verify that \texttt{xOrder} has a high sample utilization when learning an ordering adjustment.

\begin{figure}[h]
    \centering
    \includegraphics[width=.9\columnwidth]{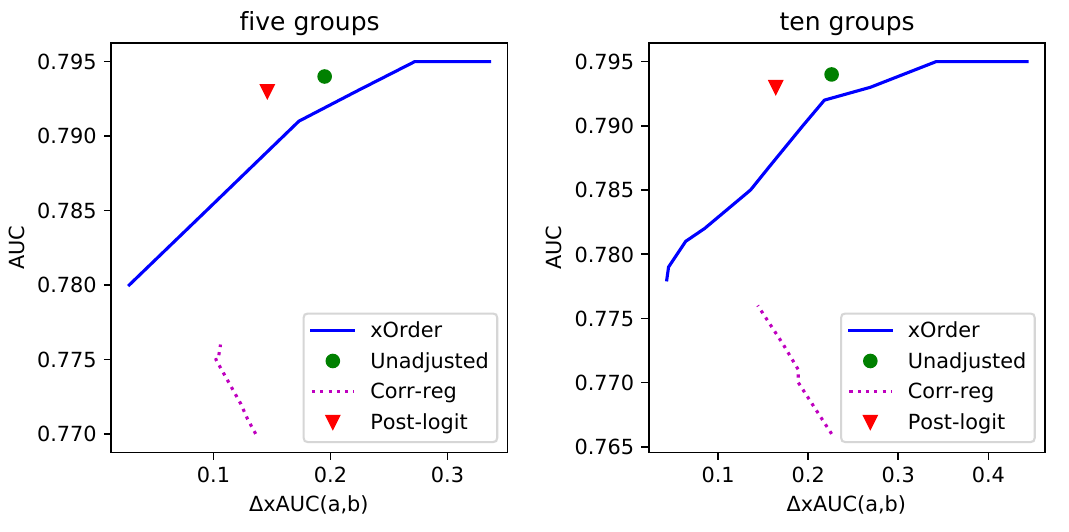}
    \caption{Experimental results on large eICU with the different number of groups (predicting the in-hospital mortality and the sensitive attribute is age).}
    \label{fig:diff_groups}
\end{figure}

From Section~\ref{sec:3-groups}, \texttt{iterative xOrder} could achieve a fair ranking when there are multiple protected groups without causing extra computation overhead. We conduct experiments to predict the in-hospital mortality on large eICU with 5 and 10 protected groups with the sensitive attribute being age\footnote{The 5 protected groups are age $\leq$45, 45$<$age$\leq$55, 55$<$age$\leq$65, 65$<$age$\leq$75, age$>$75. The 10 protected groups are age $\leq$35, 35$<$age$\leq$45, 45$<$age$\leq$50, 50$<$age$\leq$55, 55$<$age$\leq$60, 65$<$age$\leq$65, 65$<$age$\leq$70, 70$<$age$\leq$75, 75$<$age$\leq$85, age$>$85.}.

The results of all methods are shown in Figure~\ref{fig:diff_groups}. Compared with baselines, \texttt{iterative xOrder} achieves the lowest disparities and maintains a comparable utility as shown in Figure~\ref{fig:diff_groups} . Experimental results demonstrate that \texttt{iterative xOrder} could be scaled to handle large datasets with more protected groups.}

\subsection{Visiualization of the Adjustments}
Since there could exist unfairness when two very different distributions are indistinguishable in terms of AUC~\cite{vogel2020learning}, Vogel \emph{et al.} provide a pointwise ROC-based fairness constraint that encourages the equality of two distributions. To prove that our method induces a fair ranking in terms of the pointwise fairness metric~\cite{vogel2020learning}, we present the xROC curves on Adult and COMPAS dataset using linear model in Figure~\ref{fig:xauc_vis}.

From Figure~\ref{fig:xauc_vis}(a), there is a significant distribution shift of the learned risk scores without any fairness considerations. The disparities are about 0.11 (0.935-0.829) and 0.23 (0.790-0.578) on Adult and COMPAS, respectively. We show the results of the post-processing method Post-log~\cite{kallus2019fairness} in Figure~\ref{fig:xauc_vis}(b), in which it achieves the fairness on xAUC metric (the disparities are 0.02 and 0.03 on Adult and COMPAS, respectively). However, it still has a critical distribution shift from the xROC curve in Figure~\ref{fig:xauc_vis}(b).

We display the xROC curves of our method \texttt{xOrder} in Figure~\ref{fig:xauc_vis}(c). Compared with the results of post-log in Figure~\ref{fig:xauc_vis}(b), \texttt{xOrder} achieves a similar ranking fairness. For the utility, \texttt{xOrder} maintains a relative higher utility, e.g., $\mathrm{xAUC}(a,b) = \mathrm{Pr}(\mathrm{S}^{a}_{1} > \mathrm{S}^{b}_{0}) = 0.896 > 0.879$ on Adult dataset. More importantly, from the xROC curves in Figure~\ref{fig:xauc_vis}(c), our method motivates the equality of the score distributions, which decreases the disparities measured by AUC-based and pointwise ROC-based~\cite{vogel2020learning} criteria simultaneously.

\begin{figure}[htbp]
    \subfigure[Adult data set given fewer training samples(20000, 15000, 10000, 5000)]{
        \includegraphics[width=1.0\columnwidth]{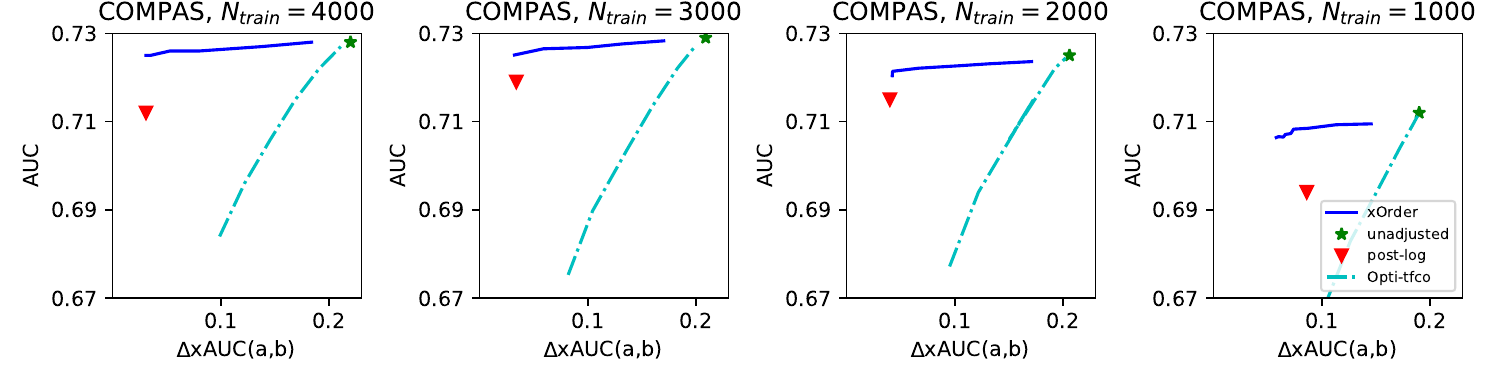}
        \label{fig:num_train_compas}
    }
    \subfigure[COMPAS data set given fewer training samples(4000, 3000, 2000, 1000)]{
        \includegraphics[width=1.0\columnwidth]{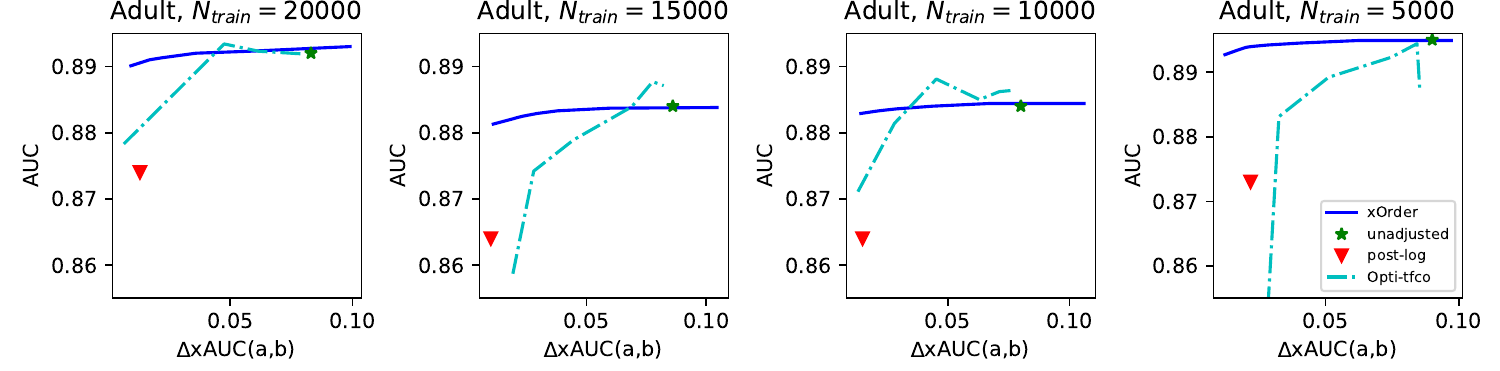}
        \label{fig:num_train_adult}
    }
    \caption{$\mathrm{AUC}$-$\mathrm{\Delta xAUC}$ trade-offs using linear model optimizing AUC with fewer training samples on Adult and COMPAS}
    \label{fig:num_train}
\end{figure}
\begin{figure}[h]
    \centering
    \subfigure[]{
        \centering
        \includegraphics[width=.46\columnwidth]{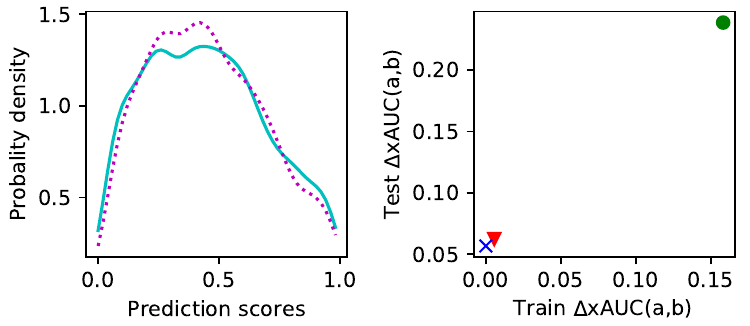}
    }%
    \subfigure[]{
        \centering
        \includegraphics[width=.46\columnwidth]{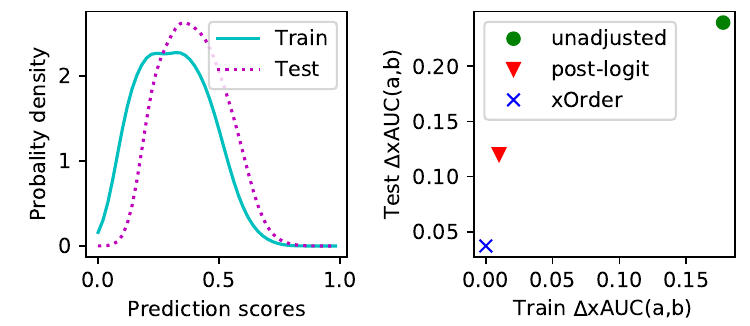}
    }%
    \caption{Result analysis on COMPAS data set with $\Delta \mathrm{xAUC}$ metric. (a) illustrates the result with a linear model. (b) illustrates the result with a bipartite rankboost model. The left part of each sub-figure is the distribution of prediction scores on training and test data, and the right part of each sub-figure plots $\Delta \mathrm{xAUC}$ on training data test data.}
    \label{fig:ana_example}
\end{figure}
\begin{figure}[h!]
    \centering
    \subfigure[]{
        \centering
        \includegraphics[width=.46\columnwidth]{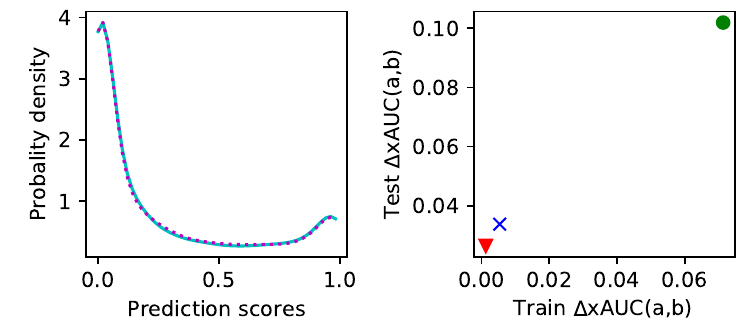}
    }%
    \subfigure[]{
        \centering
        \includegraphics[width=.46\columnwidth]{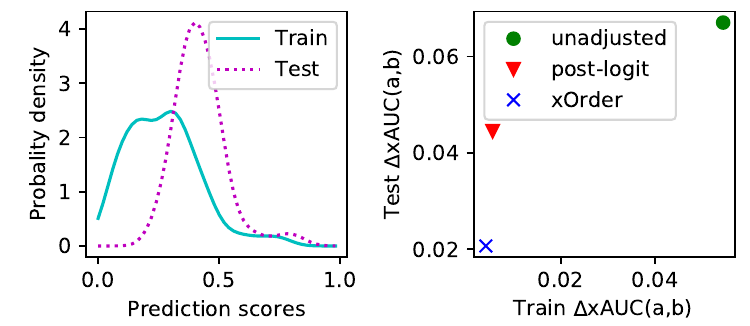}
    }%
    \caption{Result analysis on Adult data set with $\Delta \mathrm{xAUC}$ metric. (a) illustrates the result with a linear model. (b) illustrates the result with a bipartite rankboost model. The left part of each sub-figure is the distribution of prediction scores on training and test data, and the right part of each sub-figure plots $\Delta \mathrm{xAUC}$ on training data and test data.}
    \label{fig:ana_example_adult}
\end{figure}
\section{Robustness Discussion}
\subsection{Fewer Training Samples}
To assess the performance of our algorithm given fewer training samples, we conduct the experiments with a different number of training samples, in which we select a linear model as the base model with the optimization framework in \cite{narasimhan2020pairwise}. From Figure \ref{fig:num_train_compas}, as the number of training samples changed from 4000 to 1000, \texttt{xOrder} achieves a lower disparity compared to baselines. Meanwhile, \texttt{xOrder} still achieves a lower disparity while maintaining a maximum algorithm utility. Experiment results on the Adult data set shown in Figure \ref{fig:num_train_adult} also confirm the statement. As the number of training samples reduced to 5000, \texttt{xOrder} realizes the lowest disparity($\Delta \mathrm{xAUC} = 0.01$) compared to baselines($\Delta \mathrm{xAUC} = 0.02$). From the two series experiments on COMPAS and Adult, \texttt{xOrder} has stable performances on all experiments which maintain its advantages that realize a maximum algorithm utility and a minimum ranking disparity.

\subsection{Distribution Shifts}
\label{sec:6}
As a post-processing algorithm, \texttt{xOrder} adjusts the ordering directly while post-logit transforms the ranking scores to reduce the disparity. To compare the robustness of the two algorithms when faced with the difference between training and test ranking score distributions, we implement experiments on two data sets with 2 base models (linear model and RankBoost) on $\Delta \mathrm{xAUC}$ metric.

According to Figure \ref{fig:rb_result}, post-logit fails to achieve $\Delta \mathrm{xAUC}$ as low as \texttt{xOrder} with bipartite rankboost model, while both methods can achieve low $\Delta \mathrm{xAUC}$ with a linear model as shown in Figure \ref{fig:lr_xauc_result}. To analyze this phenomenon, we use COMPAS and Adult as examples in Figure \ref{fig:ana_example} and \ref{fig:ana_example_adult}. For different models, we illustrate the distributions of prediction scores $\operatorname{S}$ on training and test data. We further plot $\Delta \mathrm{xAUC}$ on training data versus $ \Delta \mathrm{xAUC}$ on test data. With a linear model, the distributions of $\operatorname{S}$ on training and test data are close to each other. In this situation, the transform relations learned from post-logit and \texttt{xOrder} can both obtain results with low $\Delta \mathrm{xAUC}$ on test data. While the distributions of the scores $\operatorname{S}$ on training and test data become significantly different, the function learned from post-logit may not be generalized well on test data to achieve low $\Delta \mathrm{xAUC}$. Similar results can be observed on the Adult data set in Figure \ref{fig:ana_example_adult}. These phenomena occur in repeat experiments on both data sets. We guess the reason is that \texttt{xOrder} adjusts the relative ordering and it is more flexible than post-logit which optimizes a logistic class function. According to the experiments, \texttt{xOrder} is more robust to such distribution difference.
\section{Conclusion}
\label{sec:7}
\revise{This paper explores the problem of ranking fairness, which has garnered considerable attention in various domains, including criminal justice, healthcare insurance plan enrollment, and job recruitment. Biased ranking models can have disastrous consequences in these areas. Therefore, it is essential to develop algorithms that can produce fair rankings. To address this issue, we propose a general post-processing framework called \texttt{xOrder}, which achieves a balance between ranking fairness and model utility, enhancing its generalization ability and trustworthiness across multiple scenarios.}
\ifCLASSOPTIONcompsoc
  \section*{Acknowledgments}
\else
  \section*{Acknowledgment}
\fi
Sen Cui, Weishen Pan and Changshui Zhang would like to acknowledge the funding by the National Key Research and Development Program (2020AAA0107800). Fei Wang would like to acknowledge the support from Amazon Web Service (AWS) Machine Learning for Research Award and Google Faculty Research Award. We would also like to appreciate the kind and favorable suggestions from Changming Xiao and Qi Yang, who are pursuing the Ph.D. degree in Tsinghua University, advised by Prof. Changshui Zhang.
\ifCLASSOPTIONcaptionsoff
  \newpage
\fi
\bibliographystyle{IEEEtran}
\bibliography{IEEEabrv,IEEEtran}
\begin{IEEEbiography}[{\includegraphics[width=1in,height=1.25in,clip,keepaspectratio]{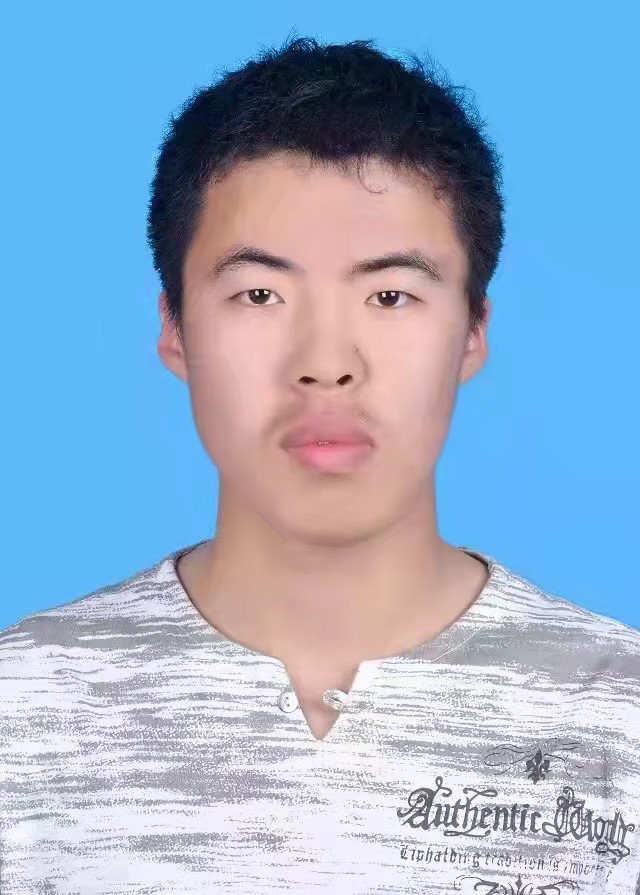}}]{Sen Cui} received his B.S. degree in the school of aerospace from Tsinghua University, Beijing, China, in 2019. He is currently pursuing the Ph.D. degree in the Department of Automation, Tsinghua University, advised by Prof. Changshui Zhang. He was a visiting student at U.C. Berkeley, Berkeley, CA, USA, in 2018, supervised by Prof. Masayoshi Tomizuka. His research interests include trustworthy AI and collaborative learning.
\end{IEEEbiography}

\begin{IEEEbiography}[{\includegraphics[width=1in,height=1.25in,clip,keepaspectratio]{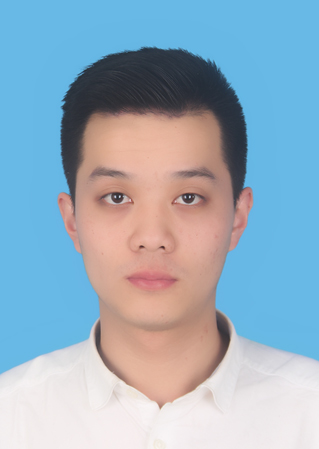}}]{Weishen Pan} received the B.S. degree from the Department of Automation, Tsinghua University, Beijing, China, in 2015. He is currently working toward the Ph.D. degree in the Department of Automation, Tsinghua University, advised by Changshui Zhang. He was a visiting scholar at Cornell University, New York, USA, in 2019. His research interests include deep learning, computer vision, and interpretable machine learning.
\end{IEEEbiography}

\begin{IEEEbiography}[{\includegraphics[width=1in,height=1.25in,clip,keepaspectratio]{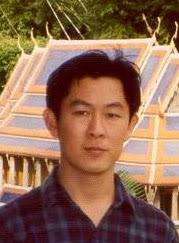}}]{Changshui Zhang} (M'02-SM'15-F'18) received the B.S. degree in mathematics from Peking University, Beijing, China, in 1986, and the M.S. and Ph.D. degrees in control science and engineering from Tsinghua University, Beijing, in 1989 and 1992, respectively. In 1992, he joined the Department of Automation, Tsinghua University, where he is currently a Professor. He is a fellow of IEEE and a member of the editorial boards of the journal IEEE Transactions on Pattern Analysis and Machine Intelligence. His current research interests include pattern recognition and machine learning.
\end{IEEEbiography}

\begin{IEEEbiography}[{\includegraphics[width=1in,height=1.25in,clip,keepaspectratio]{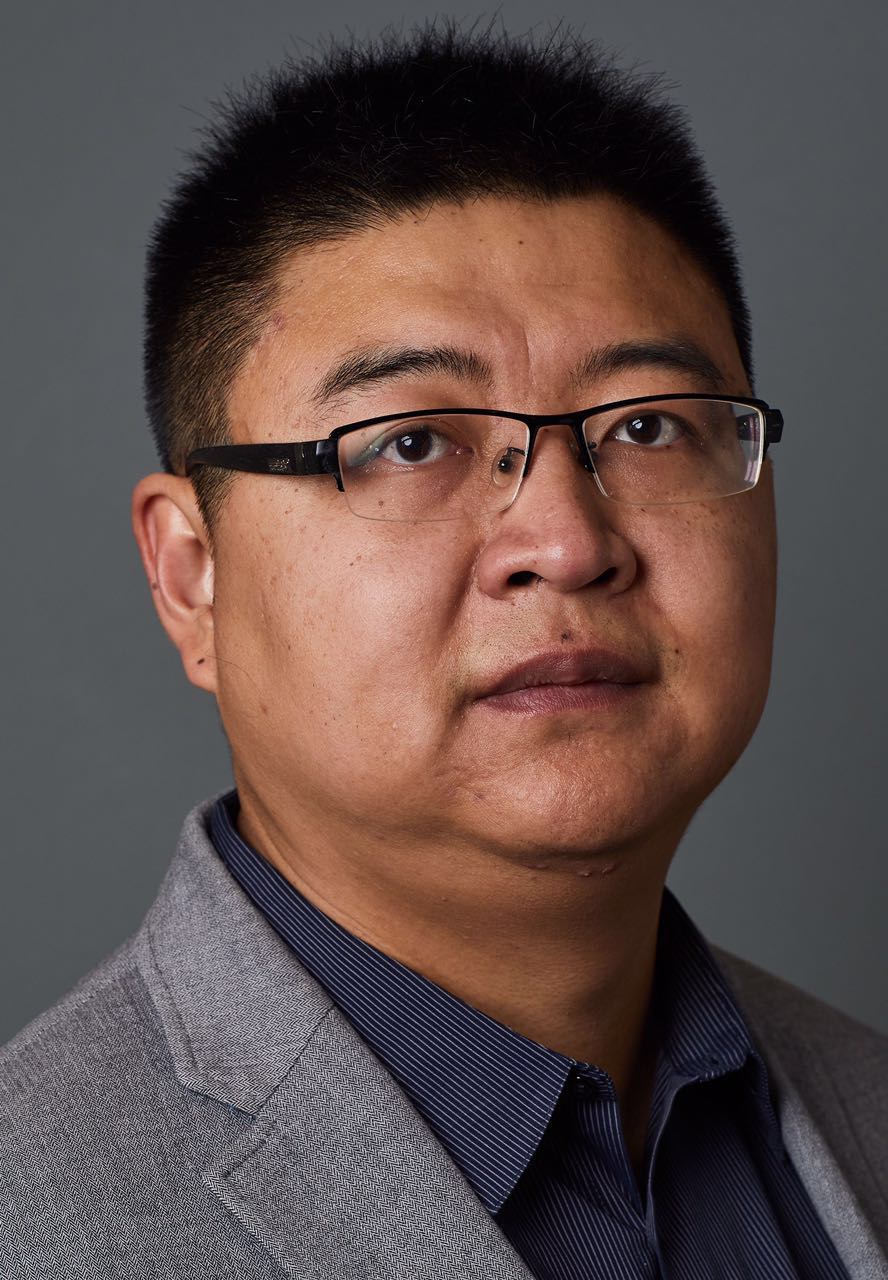}}]{Fei Wang} is an associate professor in the Division of Health Informatics of the Department of Population Health Sciences at Cornell University. He is an active editor of the journal Data Mining and Knowledge Discovery, an associate editor of the Journal of Health Informatics Research and Smart Health, and a member of the editorial boards of Pattern Recognition and the International Journal of Big Data and Analytics in Healthcare. He is a fellow of AMIA. His major research interest lies in data analytics and its applications in health informatics.
\end{IEEEbiography}

\clearpage
\includepdf[pages=-]{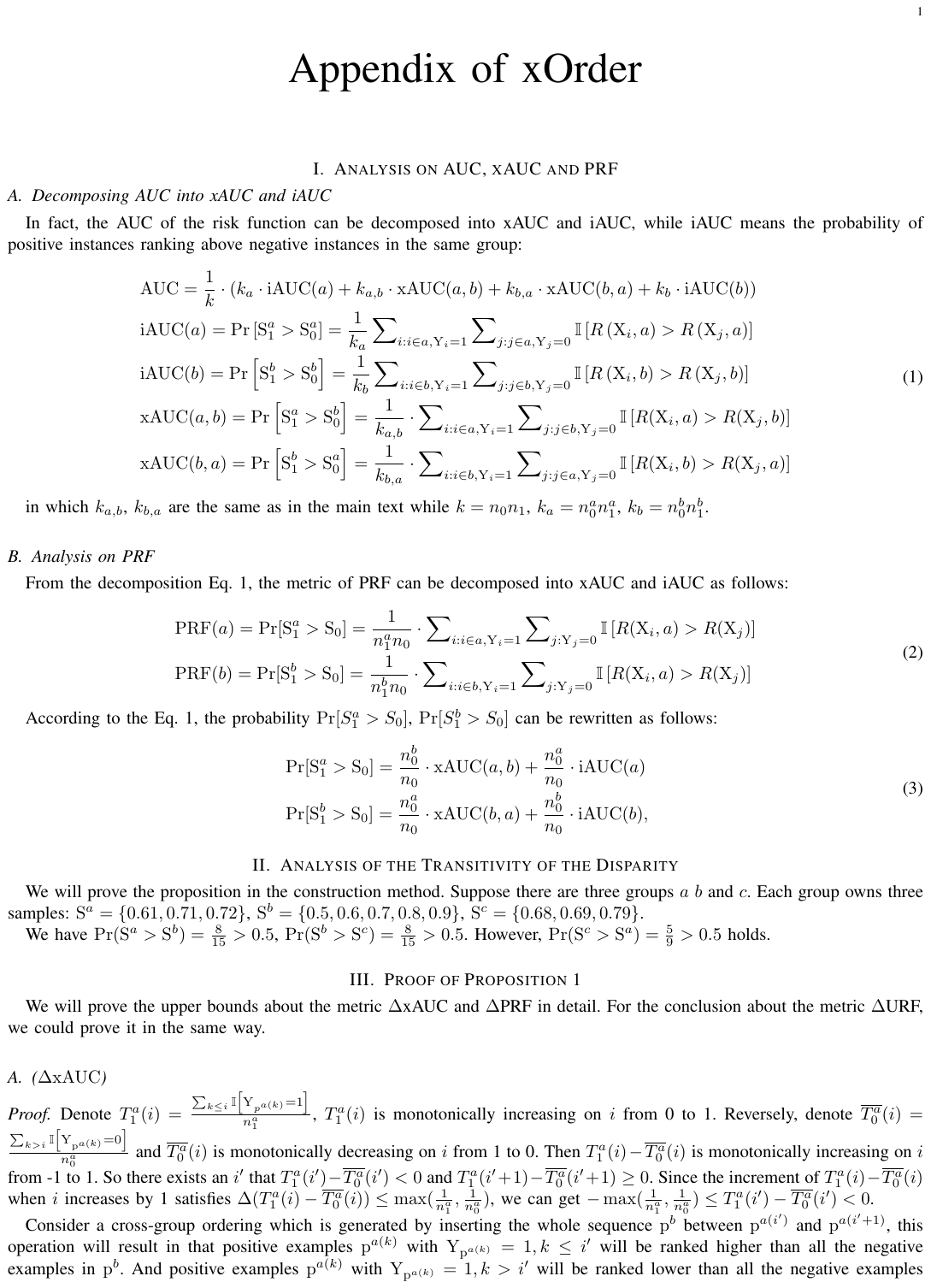}

\end{document}